\documentclass[10pt,journal]{IEEEtran}
\usepackage{amssymb}
\usepackage{graphicx}
\usepackage{picinpar}
\usepackage{booktabs}
\usepackage{float}
\usepackage{amssymb}
\usepackage{url}
\usepackage{cite}
\usepackage{mdwlist}
\usepackage{tikz}
\usepackage{amsfonts}
\usepackage{mathrsfs}
\usepackage{amssymb}
\usepackage{amsmath}
\usepackage{dsfont}
\usepackage{multirow}
\usepackage{color}
\usepackage{graphicx}
\usepackage{subfigure}
\usepackage{subfloat}
\usepackage{epstopdf}
\usepackage{booktabs, tabularx}
\usepackage{array}
\usepackage{slashbox}
\usepackage{algorithmic}
\usepackage[ruled,vlined,linesnumbered]{algorithm2e}

\def\BibTeX{{\rm B\kern-.05em{\sc i\kern-.025em b}\kern-.08em
    T\kern-.1667em\lower.7ex\hbox{E}\kern-.125emX}}

\ifCLASSINFOpdf

\fi
\begin{document}
%
\title{Two Dimensional Stochastic Configuration Networks for Image Data Analytics} 

\author{Ming Li, ~\IEEEmembership{Member,~IEEE}, Dianhui~Wang*,~\IEEEmembership{Senior Member,~IEEE}

\thanks{M. Li is with the School of Information Technology in Education, South China Normal University, Guangzhou, China, and also with the Department of Computer Science and Information
Technology, La Trobe University, Melbourne, VIC 3086, Australia. (e-mail: ming.li.ltu@gmail.com).}
\thanks{D. Wang is with the Department of Computer Science and Information
Technology, La Trobe University, Melbourne, VIC 3086, Australia, and is also with the State Key Laboratory of Synthetical Automation for Process Industries, Northeastern University,
Shenyang, Liaoning 110819, China. (e-mail: dh.wang@latrobe.edu.au).}
\thanks{*~ Corresponding author}}
\markboth{Submitted to IEEE Transactions on Cybernetics}%
{\MakeLowercase{\textit{et al.}}: }
\maketitle

\begin{abstract}
Stochastic configuration networks (SCNs) as a class of randomized learner model have been successfully employed in data analytics due to its universal approximation capability and fast modelling property. The technical essence lies in stochastically configuring hidden nodes (or basis functions) based on a supervisory mechanism rather than data-independent randomization as usually adopted for building randomized neural networks. Given image data modelling tasks, the use of one-dimensional SCNs potentially demolishes the spatial  information of images, and may result in undesirable performance. This paper extends the original SCNs to two-dimensional version, termed  2DSCNs, for  fast building randomized learners with matrix-inputs. Some theoretical analyses on the goodness of 2DSCNs against SCNs, including  the complexity of the random parameter space, and the superiority of generalization, are presented. Empirical results over one regression, four benchmark handwritten digits classification, and two human face recognition datasets demonstrate that the proposed 2DSCNs perform favourably and show good potential for image data analytics. 

\end{abstract}


\IEEEpeerreviewmaketitle

\section{Introduction}
Along with the rising wave of deep learning, neural networks, by means of their universal approximation capability and well-developed learning techniques, have achieved great success in data analytics \cite{Goodfellow-et-al-2016}. Usually, the input layer of a fully connected neural network (FCNN) is fed with vector inputs, rather than two-dimensional matrices such as images or higher dimensional tensors like videos or light fields \cite{Qi2013a, Qi2013b, Qi2018}. Technically, the vectorization operation makes the dot product (between the inputs and hidden weights) computationally feasible but inevitably induces two drawbacks: (i) the dimensionality curse issue when the number of training samples is limited; (ii) the loss of spatial information of the original muti-dimensional input. Although convolutional neural networks (CNNs) have brought about some breakthroughs in image data modelling, by means of their good potential in abstract feature extraction, power in local connectivity and parameter sharing, etc. \cite{LeCun1990}, the development of FCNNs with matrix inputs (or multidimensional inputs in general) 
\begin{figure}[!h]
\centering
\includegraphics[width=0.46\textwidth]{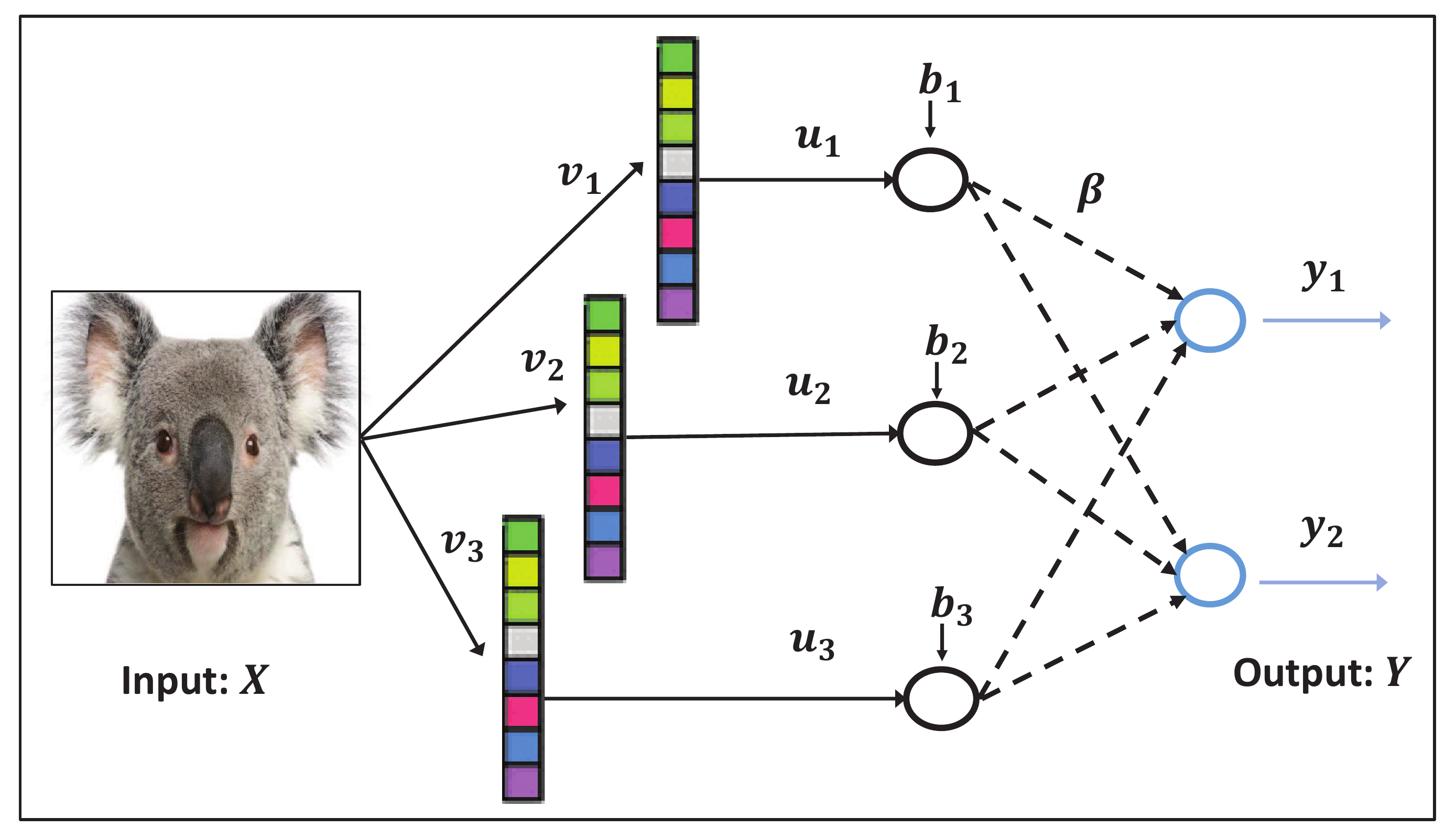}
\caption{A Typical 2D Stochastic Configuration Network  with three hidden nodes and two output nodes. The output weights (trainable) are shown as black dashed lines, whilst hidden weights and
biases (stochastically configured) are shown as fixed lines.
}\label{demo}
\end{figure}
is of great importance in terms of both theoretical and algorithmic viewpoints. In \cite{Gao2017a}, Gao et al. first nominated the term `Matrix Neural Networks (MatNet)' and extended the conventional back-prorogation (BP) algorithm \cite{Rumelhart1986} to a general version that is capable for dealing with 2D inputs. Empirical results in \cite{Gao2017a} and their parallel work \cite{Gao2017b} demonstrate some advantages of MatNet for image data modelling. Obviously, MatNet may still suffer some intrinsic drawbacks of gradient descent-based approaches such as local minimum and low convergence rate. That indicates a urgent demand for developing fast learning techniques to build FCNNs with 2D inputs, as an immediate motivation of this work.

Randomized learning techniques have demonstrated their great potential in fast building neural network models and algorithms with less computational cost \cite{Scardapane2017}. In particular, Random Vector Functional-Link (RVFL) networks developed in the early 90s \cite{Pao1994,Igelnik1995} and Stochastic Configuration Networks (SCNs) proposed recently \cite{WangandLi-SCN} are two representatives of the randomized learner models.
Technically, RVFL networks randomly assign the input weights and biases from a fixed distribution (range) that is totally data-independent, and optimize merely the output weights by solving a linear least squares problem. This trivial idea sounds computationally efficient, however, the obtained learner models may not have universal approximation capability when an inappropriate distribution (range) is used for the random assignment. This drawback indeed makes RVFL networks less practical in data modelling problems because more human intervention and/or empirical knowledge is required for problem-solving. Fortunately, SCNs, as state-of-the art randomized leaner models, were developed with rigorous theoretical fundamentals and advanced algorithm implementations \cite{WangandLi-SCN}. The success of SCNs and their extensions \cite{wang2017robust,WangandLi-DeepSCN} in fast building universal approximators with randomness has been extensively demonstrated on data analytics. Generally, the very heart of SCNs framework lies in the supervisory (data-dependent) mechanism used to stochastically (and incrementally) configure the input weights and biases from an appropriate `support range'. In the presence of multi-dimensional especially 2D inputs (e.g. images), both RVFL networks and SCNs require a regular vectorization operation before feeding the given input signal into the neural network model. Authors in \cite{Lu2014} made a first attempt on two dimensional randomized learner models, via developing RVFL networks with matrix inputs (termed 2DRVFL) with applications in image data modelling. Although some advantages of the 2D model are experimentally demonstrated, the concerned methodology/framework still suffers from the drawbacks of RVFL networks as highlighted above.

This paper develops two dimensional SCNs on the basis of our previous SCN framework \cite{WangandLi-SCN}, aiming to fast build 2D randomized learner models that are capable for resolving data anlytics with matrix inputs. We first provide a detailed algorithmic implementation for 2DSCN, followed by a convergence analysis with special reference to the universal approximation theorem of SCNs.
Then, some technical differences between 2DSCN and SCN are presented with highlights in various aspects, such as the support range for random parameters, the complexity for parameter space, data structure preservation. Among that and interestingly, our work is the first to think about a potential relationship between 2DSCN and CNN in problem-solving, that is, computations involved in 2DSCN in some sense can  be viewed as equivalent to the `convolution' and `pooling' tricks performed in CNN structure. Later, some technical issues around why randomized learner models produced by 2DSCN algorithm are more prone to have a better generalization ability are investigated in-depth. In particular, some solid results from statistical learning theory are revisited with our special interpretation, for the purpose of qualitative analysis on learner models' generalization power and useful insights on certain very influential factors. Besides, we provide an intuitive sense that 2DSCN may exhibit similar philosophy as concerned in DropConnect \cite{Wan2013} for effectively alleviating over-fitting. Importantly, to make a reasonable and practical judgement on the generalization ability of an obtained randomized learner model, we make efforts towards developing a nearly sharp estimation about the model's test error upper bound, thereby one can effectively predict the generalization performance. Extensive experimental study on both regression and classification problems (with matrix inputs setup) have demonstrated remarkable advantages of 2DSCN on image data modelling, compared to some existing randomized learning techniques. Also, our theoretical analysis has been successfully verified by the statistical simulation results. Overall, our main contributions can be summarized in three-fold:
   \begin{itemize}
     \item From the algorithmic perspective, we extend our original SCN framework to 2D version, and the proposed 2DSCN algorithm can effectively deal with data modelling tasks with matrix inputs, compared with some existing randomized learning techniques;
     \item Theoretically, the universal approximation property of 2DSCN-based learner models is verified and some technical differences between 1D and 2D randomized learner models are investigated in terms of various perspectives. Importantly, we provide an upper bound for the test error of a given randomized learner model, and demonstrate in theory how the hidden layer output matrix (computationally associated with the training inputs, the hidden input weights and biases, the number of hidden nodes, etc.) and the output weights can affect the randomized learner model's generalization power.
     \item For practical applications, the merits of the developed 2DSCN algorithm on image data analytics have been illustrated on various benchmark tasks, such as the rotation angles predication for handwritten digits, handwritten digits classification, and human face recognition. The extensive experimental study conducted in this paper can lend some empirical support to end-users who would like to employ FCNNs rather than CNNs in image data modeling.
   \end{itemize}

The remainder of this paper is organized as follows. Section II provides some related work including the 2D random vector functional link (RVFL) networks and our original SCN framework. Section III details the proposed 2D stochastic configuration networks (2DSCNs) with algorithmic description, technical highlights, and some theoretical explanation, aiming to distinguish 2DSCN from the other randomized learning techniques. Section IV presents experimental study in terms of both image-based regression and classification problems, and Section V concludes this paper with further remarks and expectation.
\section{Related Work}
This section reviews two types of randomized learner models with highlights in their technical discrepancy. First, 2DRVFL networks as an extension of RVFL networks can deal with matrix inputs but could not guarantee universal approximation in data modelling, which causes some infeasibility as well as uncertainty for problem-solving; SCNs as an advanced universal approximator have demonstrated their effectiveness and efficiency in data analytics with vector inputs as usually done. Basically, the brief examination of these two methodology motivates us to think about the formulation of 2DSCN and also its potential advantages in image data modelling problems, as to be detailed in the following section.
\subsection{2DRVFL Networks}
2DRVFL networks with matrix inputs has been empirically studied in \cite{Lu2014}. Technically, it can be viewed as a trivial extension of the original RVFL networks in computation via employing two sets of input weights acting as matrix transformation over the left and right sides of inputs. Here we start directly with the problem formulation for 2DRVFL, rather than revisit the basics of RVFL networks.
Given $N$ training instances $(x_i,t_i)$ sampled from an unknown function, with inputs $x_i\in \mathds{R}^{d_1\times d_2}$, outputs $t_i\in \mathds{R}^m$, training a 2DRVFL learner model with $L$ hidden nodes
is equivalent to solving a linear least squares (LS) problem ($w.r.t$ the output weights), i.e,
\begin{equation*}
  \min_{\beta_1,\ldots,\beta_j}\sum_{i=1}^{N}\|\sum_{j=1}^{L}\beta_{j}\phi(u^{\mathrm{T}}_jx_iv_j+b_j)-t_i\|^2,
\end{equation*}
where $u_j$, $v_j$, $b_j$ are randomly assigned from $[-\lambda,\lambda]^{d_1}$, $[-\lambda,\lambda]^{d_2}$, $[-\lambda,\lambda]$, respectively and remain fixed. $g(\cdot)$ is the activation function.

The above LS problem can be represented by a matrix form, i.e.,
\begin{equation}\label{ls}
\beta^{*}=\arg \min_{\beta}\|H\beta-T\|_F^2
\end{equation}
where
\begin{equation*}
H=\left(
\begin{array}{ccc}
g(u^{\mathrm{T}}_1x_{1}v_{1}+b_{1}) & \cdots & g(u^{\mathrm{T}}_Lx_{1}v_{L}+b_{L}) \\
\vdots & \cdots & \vdots\\
g(u^{\mathrm{T}}_1x_{N}v_{1}+b_{1})  & \cdots & g(u^{\mathrm{T}}_Lx_{N}v_{L}+b_{L})
\end{array}\right)
\end{equation*}
is the hidden layer output matrix, $T=[t_1,t_2,\ldots,t_N]^{\mathrm{T}}$, $\beta=[\beta_1,\beta_2,\ldots,\beta_L]^{\mathrm{T}}$. A closed form solution can be obtained by using the pseudo-inverse method, i.e., $\beta^{*}=H^{\dagger}T$.

\textbf{Remark 1.} Although RVFL networks (with either vector or matrix inputs) allows fast building a model by randomly assigning input weights and biases, some key technical issues are still unresolved. Theoretically, approximation error for this kind of randomized learner model are bounded in the statistical sense, which means preferable approximation performance is not guaranteed for every random assignment of the hidden parameters \cite{Igelnik1995}. Besides, it has been proved that in the absence of such additional conditions, one may observe exponential growth of the number of terms needed to approximate a non-linear map, and/or the resulting learner model will be extreme sensitivity to the parameters \cite{SI-Gorban2016}. From the algorithmic perspective, all these theoretical predictions do not address the learning algorithm or implementation issues for the randomized learner. Practical usage of this kind of randomized model encounter one key technical difficulty, that is, how to find an appropriate range for randomly assigning hidden parameters with considerable confidence to ensure the universal approximation property. So far, the most accurate (and trivial) way for implementing RVFL networks should employ trial-and-error/rule-of-thumb for parameter setting, that is to say, one needs perform various setting of $\lambda$ before getting an acceptable learner model.
This trick sounds practical but still has potential drawbacks due to uncertainty causes by the randomness, as theoretically and empirically studied in \cite{LiandWang2016}. We also note that one can try out different random selection range for $u_j$ and $v_j$ in 2DRVFL, such as $u_j\in [-\lambda_1,\lambda_1]^{d_1}$ and $v_j\in [-\lambda_2,\lambda_2]^{d_2}$, but may need more grid-searching in algorithm implementation to find out the 'best' collection $\{\lambda_1^{*},\lambda_2^{*}\}$.

\subsection{SCN framework}
Our recent work \cite{WangandLi-SCN} is the first to touch the foundation of building universal approximator with random basis functions. More precisely, a new type of randomized learner model, termed stochastic configuration networks (SCNs), is developed by implanting a `data-dependent' supervisory mechanism to the random assignment of input weights and biases. Readers who are interested in a complete roadmap of this novel work can refer to \cite{WangandLi-SCN}. Here we just briefly revisit the essence and highlight some technical points.

Let $\Gamma:=\{g_1, g_2, g_3...\}$ represent a set of real-valued functions, and span$(\Gamma)$ stands for the associated function space spanned by $\Gamma$. $L_{2}(K)$ denote the space of all Lebesgue measurable functions $f=[f_1,f_2,\ldots,f_m]:\mathds{R}^{d}\rightarrow \mathds{R}^{m}$ defined on $K\subset \mathds{R}^{d}$, with the $L_2$ norm defined as
\begin{equation}\label{multiple_lp}
  \|f\|:=\left(\sum_{q=1}^{m}\int_{D}|f_q(x)|^2dx\right)^{1/2}<\infty.
\end{equation}
Given another vector-valued function $\phi=[\phi_1,\phi_2,\ldots,\phi_m]:\mathds{R}^{d}\rightarrow \mathds{R}^{m}$, the inner product of $\phi$ and $f$ is defined as
\begin{equation}\label{multiple_inner}
  \langle f,\phi\rangle:=\sum_{q=1}^{m}\langle f_q,\phi_q\rangle=\sum_{q=1}^{m}\int_{K}f_q(x)\phi_q(x)dx.
\end{equation}
Note that this definition becomes the trivial case when $m=1$, corresponding to a real-valued function defined on a compact set.

Before revising the universal approximation theory behind SCNs, we recall the problem formulation as follows. For a target function $f:\mathds{R}^{d}\rightarrow \mathds{R}^{m}$, suppose that we have already built a neural network learner model with only one hidden layer and $L-1$ hidden nodes, i.e, $f_{L-1}(x)=\sum_{j=1}^{L-1}\beta_jg_j(w_j^\mathrm{T}x+b_j)$ ($L=1,2,\ldots$, $f_0=0$), with $\beta_j=[\beta_{j,1},\ldots,\beta_{j,m}]^\mathrm{T}$, and residual error $e_{L-1}=f-f_{L-1}=[e_{L-1,1},\ldots,e_{L-1,m}]$ far away from an acceptable accuracy level, our SCN framework can successfully offer a fast solution to incrementally add $\beta_L$, $g_L$ ($w_L$ and $b_L$) leading to $f_{L}=f_{L-1}+\beta_Lg_L$ until the residual error $e_L=f-f_L$ falls into an expected tolerance $\epsilon$. The following Theorem 1 restates the universal approximation property of SCNs, corresponding to Theorem 7 in \cite{WangandLi-SCN}.

\textbf{Theorem 1.} Suppose that span($\Gamma$) is dense in $L_2$ space and $\forall g\in \Gamma$, $0<\|g\|<b_g$ for some $b_g\in \mathds{R}^{+}$. Given $0<r<1$ and a nonnegative real number sequence $\{\mu_L\}$ with $\lim_{L\rightarrow+\infty}\mu_L=0$, $\mu_L\leq (1-r)$, for $L=1,2,\ldots$, denoted by
\begin{equation}
\delta_{L}=\sum_{q=1}^{m}\delta_{L,q}, \delta_{L,q}=(1-r-\mu_L)\|e_{L-1,q}\|^2, q=1,2,\ldots,m,
\end{equation}
if the random basis function $g_L$ is generated to satisfy the following inequalities:
\begin{equation}\label{step3}
\langle e_{L-1,q},g_L\rangle^2\geq b_g^2\delta_{L,q}, q=1,2,\ldots,m,
\end{equation}
and the output weights are evaluated by
\begin{equation}\label{step4}
[\beta_1, \beta_2,\ldots,\beta_{L}]=\arg \min_{\beta}\|f-\sum_{j=1}^{L}\beta_jg_j\|,
\end{equation}
it holds that $\lim_{L\rightarrow +\infty}\|f-f_L\|=0,$ where $f_L=\sum_{j=1}^{L}\beta_{j}g_j$, $\beta_{j}=[\beta_{j,1},\ldots,\beta_{j,m}]^{\mathrm{T}}$.

Basically, the algorithmic procedures for building SCNs can be summarized as repeating the following sessions with $L=1,2,3\ldots.$ until the given training error tolerance is reached:
\begin{itemize}
  \item Stochastically configure a new hidden node $g_L$ (i.e, find out random $w_L$ and $b_L$ from support range) based on the inequality (\ref{step3});
  \item Evaluate $\beta$ by solving the linear least squares problem expressed in Eq. (\ref{step4});
  \item Calculate the current training error $e_L$ and check the termination condition is met or not.
\end{itemize}

\textbf{Remark 2.} We would like to highlight that SCNs outperforms some existing randomized learning techniques (e.g. RVFL networks) that employ a totally data-independent randomization in training process, and demonstrate considerable advantages in building fast learner models with sound learning and generalization ability. It implies a good potential for dealing with online stream and/or big data analytics. Recently, some extensions of SCNs
are proposed towards various viewpoints. In \cite{wang2017stochastic}, an ensemble version of SCNs with heterogeneous features was developed with applications in large-scale data analytics. In \cite{WangandLi-DeepSCN}, we have generalized our SCNs to a deep version, termed as DeepSCNs, with both theoretical analysis and algorithm implementation. It has been empirically illustrated that DeepSCNs can be constructed efficiently (much faster than other deep neural networks) and share many great features, such as learning representation and consistency property between learning and generalization. Besides in \cite{wang2017robust}, we built robust SCNs for the purpose of uncertain data modelling. This series of work to some extent exhibits the effectiveness of SCN framework and have displayed an advisable and useful way on studying/implantng randomness in neural networks.

\section{2D Stochastic Configuration Networks}
This section details our proposal for two dimensional stochastic configuration networks (2DSCN). First, based on our original SCN framework, we can straightforwardly present the algorithm description for 2DSCN, followed by theoretically verifying the convergence property. Then, comparison around some technical points between these two methods are discussed. Afterwards, a theoretical analysis why randomized learner models with 2D inputs have a good potential for inducing better generalization is provided.
\subsection{Algorithm Implementation}
On the basis of SCN framework, the problem of building 2DSCN can be formulated as follows. Given a target function $f:\mathds{R}^{d_1\times d_2}\rightarrow \mathds{R}^{m}$, suppose that a 2DSCN with $L-1$ hidden nodes has already been constructed, that is, $f_{L-1}(x)=\sum_{j=1}^{L-1}\beta_jg_j(u_j^\mathrm{T}xv_j+b_j)$ ($L=1,2,\ldots$, $f_0=0$), where $g(\cdot)$ represents the activation function, $u_j\in \mathds{R}^{d_1}$, $v_j\in \mathds{R}^{d_1}$ stand for the collection of input weights (to be stochastically configured with certain constrains), $\beta_j=[\beta_{j,1},\ldots,\beta_{j,m}]^\mathrm{T}$ are the output weights. With the current residual error denoted by $e_{L-1}=f-f_{L-1}=[e_{L-1,1},\ldots,e_{L-1,m}]$, which as supposed does not reach a pre-defined tolerance level, our objective is to fast generate a new hidden node $g_L$ (in lieu of $u_L$, $v_L$, and $b_L$) so that the resulted model $f_{L}$ has an improved residual error after evaluating all the output weights $\beta_1,\beta_2,\ldots,\beta_L$ based on a linear least squares problem.

Suppose we have a training dataset with inputs $X=\{x_1,x_2,\ldots,x_N\}$, $x_i\in \mathds{R}^{d_1\times d_2}$ and its corresponding outputs $T=\{t_1,t_2,\ldots,t_N\}$, where $t_i=[t_{i,1},\ldots,t_{i,m}]^\mathrm{T}\in \mathds{R}^{m}$, $i=1,\ldots,N$, sampled based on a target function $f:\mathds{R}^{d_1\times d_2}\rightarrow \mathds{R}^{m}$. Denoted by $e_{L-1}:=e_{L-1}(X)=[e_{L-1,1}(X),e_{L-1,2}(X),\ldots,e_{L-1,m}(X)]^\mathrm{T}\in \mathds{R}^{N\times m}$ as the corresponding residual error vector before adding the $L$-th new hidden node, where $e_{L-1,q}:=e_{L-1,q}(X)=[e_{L-1,q}(x_1),\ldots,e_{L-1,q}(x_N)]\in \mathds{R}^N$ with $q=1,2,\ldots,m$. With $N$ two dimensional inputs $\{x_1,x_2\ldots,x_N\}$, the $L$-th hidden node activation can be expressed as
\begin{equation}\label{newhidden}
h_L:=h_L(X)=[g_{L}(u_L^\mathrm{T}x_1v_L+b_L),\ldots,g_{L}(u_L^\mathrm{T}x_Nv_L+b_L)]^\mathrm{T},
\end{equation}
where $u_L\in \mathds{R}^{d_1}$ and $v_L\in \mathds{R}^{d_2}$ are input weights, $b_L$ is the bias.

Denote a set of temporal variables $\xi_{L,q}, q=1,2,...,m$ as follows:
\begin{eqnarray}\label{inequlity}
\xi_{L,q}=\frac{\Big(e^{\mathrm{T}}_{L-1,q}\:\:h_L\Big)^2}{h^{\mathrm{T}}_L h_L}-(1-r)e^{\mathrm{T}}_{L-1,q}e_{L-1,q}.
\end{eqnarray}
Based on Theorem 1, it is natural to think about the inequality constrain for building 2DSCN by letting $\sum_{q}^{m}\xi_{L,q}\geq 0$.

After successfully adding the $L$-th hidden node ($g_L$), the current hidden layer output matrix can be expressed as  $H_L=[h_1,h_2,\ldots,h_L]$.
Then, the output weights are evaluated by solving a least squares problem, i.e.,
\begin{equation}\label{output_weights}
  \beta^{*}=\arg\min_{\beta}\|H_L\beta-T\|_{F}^2=H^{\dagger}_LT,
\end{equation}
where $H^{\dagger}_L$ is the Moore-Penrose generalized inverse \cite{Lancaster1985} and $\|\cdot\|_{F}$ represents the Frobenius norm.

\begin{algorithm}[htbp!]
\footnotesize
\SetAlgoLined
\SetKwInOut{Input}{Input}\SetKwInOut{Output}{Output}
\caption{\footnotesize 2DSCN}
\label{Algorithm 2DSCN}
\DontPrintSemicolon
\Input{Training inputs $X=\{x_1,x_2,\ldots,x_N\}$, $x_i\in R^{d_1\times d_2}$, outputs $T=\{t_1,t_2,\ldots,t_N\}$, $t_i\in R^{m}$; The maximum number of hidden nodes $L_{max}$; The expected error tolerance $\epsilon$; The maximum times of random configuration $T_{max}$; Two sets of scalars $\Upsilon=\{\lambda_{1},\ldots,\lambda_{end}\}$ and $\mathcal{R}=\{r_{1},\ldots,r_{end}\}$}
\Output{A 2DSCN model}
\BlankLine
\textbf{Initialization}: $e_0:=[t_1^\mathrm{T},t_2^\mathrm{T},\ldots,t_N^\mathrm{T}]^\mathrm{T}$, $\Omega,W:=[\:\:]$;\\

 \While{$L\leq L_{max}$ and $\|e_0\|_{F}>\epsilon$}{
   \For{$\lambda \in \Upsilon$}{
     \For{$r\in \mathcal{R}$}{
       \For{$k=1,2\ldots,T_{max}$}{
        Randomly assign $u_L$, $v_L$, $b_L$ from $[-\lambda,\lambda]^{d_1}$, $[-\lambda,\lambda]^{d_2}$, $[-\lambda,\lambda]$, respectively;\\
        Calculate $h_L$ by Eq. (\ref{newhidden}), and $\xi_{L,q}$ by Eq. (\ref{inequlity});\\
        \eIf{$min\{\xi_{L,1},\xi_{L,2},...,\xi_{L,m}\}\geq 0$}{
        \textbf{Save} $w_L$ and $b_L$ in $W$, $\xi_L\!\!=\!\!\sum_{q=1}^{m}\!\xi_{L,q}$ in $\Omega$;}
           {go back to \textbf{Step 5};}
           }
       \eIf {$W$ is not empty}   {Find ($u_L^{*},v_L^{*},b_L^{*})$ maximizing $\xi_L$ in $\Omega$, and set the hidden output matrix $H_L=[h^*_1,h^*_2,\ldots,h^*_L]$;\\
       \textbf{Break} (go to \textbf{Step 23}); }
       {\textbf{Continue}: go to \textbf{Step 5};}
       }
   }
   Calculate $\beta^{*}=[\beta^{*}_1,\beta^{*}_2,\ldots,\beta^{*}_L]^\mathrm{T}$ based on Eq. (\ref{output_weights});\\
   Calculate $e_L=H_L\beta^{*}-T$ and renew ${e}_0:=e_L$, $L:=L+1$;
 }

\textbf{Return}:$L^{*},\beta^{*}, u^{*}, v^{*}, b^{*}$.
\end{algorithm}
\subsection{Convergence Analysis}
The key to verify the convergence of Algorithm 1 is to analyze the universal approximation property of 2DSCN. Recall the proof of Theorem 1 (Theorem 7 in \cite{WangandLi-SCN}), one can observe that it is the inequality constrains that dominant the whole deduction, rather than the form of input weights (either vector or matrix). In fact, it still holds that $\|e_L\|$ is monotonically decreasing and convergent, $\|e_L\|^2\leq r\|e_{L-1}\|^2$ for a given $r\in (0,1)$. Therefore, $\lim_{L\rightarrow+\infty}\|e_L\|=0$.

We remark that $r$ value is varying during the whole incremental process and the same approach intuitively applies to verify the convergence. Also, it sounds logical to set $r$ as a sequence with monotonically increasing values, because it will become more difficult to meet the inequality condition after considerable amount of hidden nodes are successfully configured.  To some extent, this user-determined (and problem-dependent) parameter affects the algorithm convergence speed. In particular, one can set $r$ sequence (monotonically increasing) with initial value quite close to one, which can ease the configuration phase when adding one hidden node as the inequality condition can be easily satisfied. Alternatively, user can start with a relatively small value (but cannot be too small), which however requires more configuration trials at one single step to find suitable input weights and biases that fit the inequality condition. It can lead to huge computational burden or even more unnecessary fails during the configuration phase. Since the convergence property is guaranteed theoretically, one can think about some practical guideline for setting $r$ sequence with reference to their practical task. Based on our experience, the first trick, i.e., initializing $r$ with value close to one and then monotonically increase (progressively approaching one), offers more feasibility in algorithm implementation. Later in Section IV, we will recall this note in our experimental setup.
\subsection{Comparison with SCNs}
\subsubsection{Support Range for Random Parameters}
Technically, 2DSCN still inherits the essence of our original SCN framework, that is, stochastically configuring basis functions in light of a supervisory mechanism (see Theorem 1). This kind of data-dependent randomization way can effectively and efficiently locate the `support range', where one can randomly generate hidden nodes with insurance for building universal approximators. Despite this common character, differences between support ranges induced by these two methods should be highlighted. Computationally, it holds that
\begin{eqnarray*}\label{trace}
  u^{\mathrm{T}}xv&=&\mbox{Tr}(u^{\mathrm{T}}xv)=\mbox{Tr}(xvu^{\mathrm{T}})\nonumber\\
  &=&\mbox{Tr}((uv)^{\mathrm{T}}x)=
  (vec(uv^{\mathrm{T}}))^{\mathrm{T}}vec(x),
\end{eqnarray*}
where $\mbox{Tr}$ means the matrix trace, $u\in \mathds{R}^{d_1}$, $v\in \mathds{R}^{d_2}$, $vec(\cdot)\in \mathds{R}^{d_1d_2}$ stands for vectorization of a given 2D array. \newline
We observe that although $(vec(uv^{\mathrm{T}}))^{\mathrm{T}}vec(x)$ can be viewed as a regular dot product (between the hidden weight vector and input) computation performed in SCN, the resulted ($d_1d_2$)-dimensional vector $vec(uv^{\mathrm{T}})$ may exhibit different distribution, in contrast to a random ($d_1d_2$)-dimensional vector from SCN-induced support range.

We should also note that there is no special requirements for the initial distribution of $u$
and $v$ performed in the algorithm implementation. For instance, one can set two different range parameter sets $\Upsilon_u=\{\lambda^{u}_{1},\ldots,\lambda^{u}_{end}\}$ and $\Upsilon_v=\{\lambda^{v}_{1},\ldots,\lambda^{v}_{end}\}$ respectively in their experimental setup. If so, in algorithm design, one more loop is need for searching appropriate $\lambda^{v}$ from $\Upsilon_v$ when $\lambda^{u}$ is chosen and fixed, or vice versa. Since universal approximation capability is always guaranteed, this complex manipulation sounds not computationally efficient in practical implementation .
For simplicity, we just use the same random range setting for $u$ and $v$, i.e., merely $\Upsilon$, as noticed in step 3 of the above Algorithm 1.

In practice, $u$ or $v$, which can be viewed as row/column-direction hidden weight, has its own support range, which relies on their initially employed distribution ($\Upsilon_u$ or $\Upsilon_v$) and the inequality constrain for hidden node configuration. Regarding discrepancies between 2DSCN and SCN in random parameter distribution, we will elaborate more details at the end of this section.
\subsubsection{Parameter Space}
Despite  that neural networks can universally approximate complex multivariate functions, they still suffers from difficulties on high-dimensional problems where the number of input features is much larger than the number of observations. Generally, a huge number of training observations is required for training/building an acceptable approximator, as normally performed in deep learning community. Empirically, problems with very limited number of training samples but of very high dimension usually need further technical concerns in algorithm development, like feature selection or learning representation with sparsity (e.g, Lasso). To avoid high-dimensional inputs and seek useful input features for the alleviation of overfitting are important and essential for the majority of machine learning techniques.

It is clear that one 2DSCN model with $L$ hidden nodes has $L$ $d_1$-dimensional weights and $L$ $d_2$-dimensional input weights, $L$ biases (scalar), $L$ $m$-dimensional output weights, that is, $L\times(d_1+d_2+2)$ parameters in total; whist SCN model with the same structure has $L$ $(d_1\times d_2)$-dimensional input weights and the same amount of biases and output weights, i.e., $L\times(d_1d_2+1+m)$ parameters altogether. Technically, in SCN, it can impose a high dimensional parameter space that may cause potential difficulties to meet the stochastic configuration inequality (\ref{step3}), especially when the number of training samples is far lower than the dimensionality of the input weights. Besides, for a relatively large $L$, huge memory is needed for saving $L\times(d_1d_2+1+m)$ parameters in computation. On the contrary, 2DSCN can effectively ease the high-dimensional issue and to some extent economize physical memory in practice.
\subsubsection{Data Structure Preservation}
\begin{figure}[htbp]
\centering
\includegraphics[width=0.48\textwidth]{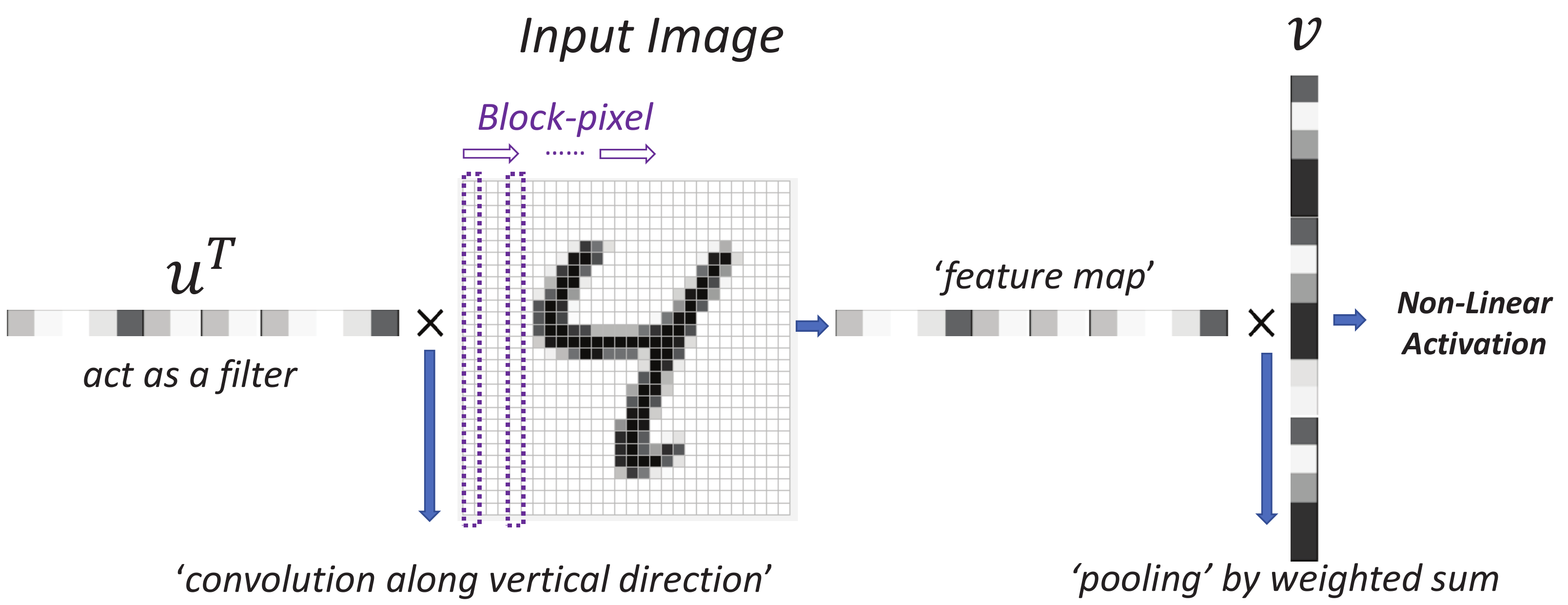}
\caption{Schematic diagram for computational equivalence between 2DSCN and CNN.}\label{schematic_2DSCN}
\end{figure}

It sounds logical that 2DSCN has some advantages in preserving the spatial details of the given input images, due to that it cares about the 2D-neighborhood information (the order in which pixels appear) of the input rather than a simple vectorization operation performed in SCN. This argument has been raised and commonly accepted in literature, however, there is no sufficient scientific evidence verifying why and how the vectorization trick affect the structural information of the 2D inputs. In this part, we aims at examining the resemblance between 2DSCN methodology and convolutional neural networks (CNNs) in terms of computational perspective. A schematic diagram is plotted in Fig. \ref{schematic_2DSCN} and corresponding explanations are as follows.

Recall Eq. (\ref{trace}), the left low-dimensional vector $u^{\mathrm{T}}$ is acting as a `filter' used in extracting some random features from the 2D input $x$. In other words, each column of $x$ is now considered as a block, i.e., image $x=[x_1,x_2,\ldots,x_{d_2}]$ is supposed to be represented by $d_2$ block-pixels, then $u^{\mathrm{T}}x$ can be viewed as a `convolution' operation between the `filter' $u^{\mathrm{T}}$ (of size $1\times d_1$) and the input $x$ along the vertical direction, leading to a feature map $[u^{\mathrm{T}}x_1,u^{\mathrm{T}}x_2,\ldots,u^{\mathrm{T}}x_{d_2}]$. Then, a `pooling' operation, conducted by calculating a weighted sum of the obtained feature map, is used to aggregate feature information.

As a conjecture, 2DSCN might have some technical merits in common with CNNs for image data analytics. More theoretical and/or empirical research on this judgment are left for our future study.
\subsection{Superiority in Generalization}
In this part, we will investigate in-depth why 2DSCN (and 2DRVFL) potentially leads to a better generalization performance than SCN (and RVFL). Four supportive theories (ST1 to ST4) are presented to explain our intuition prediction, that is, the stochastically configured input weights and biases of 2DSCN to a great extent are more prone
to result in lower generalization error. Later some statistical verification are demonstrated to further justify our theoretical interpretation.

\textbf{\emph{ST1: Learning Less-Overlapping Representations.}} Typically, elements of a weight vector have one-to-one correspondence with observed features and a weight vector is oftentimes interpreted by examining the top observed-features that correspond to the largest weights in this vector. In \cite{Xie2017arXiv}, the authors proposed a non-overlapness promoting regularization learning framework to improve interpretability and help alleviate overfitting. It imposes a structural constraint over the weight vectors, thus can effectively shrink the complexity of the function class induced by the neural network models and improve the generalization performance on unseen data. Assume that a model is parameterized by $L$ vectors $\mathcal{W}=\{\bar{w}_i\}_{i=1}^{L}$, \cite{Xie2017arXiv} proposed a hybrid regularizer consisting of a orthogonality-promoting term and sparsity-promoting term, denoted by
\begin{equation*}
  \Omega(\mathcal{W})=\mbox{tr}(M)-\log\det(M)+\gamma\sum_{i=1}^{L}\|\bar{w}_i\|_{l_1},
\end{equation*}
where $M$ is the Gram matrix associated with $\mathcal{W}$, i.e., $M_{i,j}=\bar{w}_i^{\mathrm{T}}\bar{w}_j$, $\gamma$ is a tradeoff parameter between these two regularizers.

Theoretically, the first term $\mbox{tr}(M)-\log\det(M)$ controls the level of near-orthogonal over the weight vectors from  $\mathcal{W}$, while the second term $\sum_{i=1}^{L}\|\bar{w}_i\|_{l_1}$ encourage $w_i\in \mathcal{W}$ to have more elements close to zero. It is empirically verified that this hybrid form of regularizer can contribute to learner models with better generalization performance \cite{Xie2017arXiv}.

Back to our thesis, we can roughly explain why 2D models (2DSCN and 2DRVFL) can outperform 1D models (SCN and RVFL), and simultaneously, why SCN-based models are better than RVFL-based ones: (i) Generally, there is no big difference between SCN and 2DSCN on the near-orthogonal level of $\mathcal{W}$, however, random weights in 2DSCN can have higher level of sparsity than that in SCN, hence leading to a smaller $\sum_{i=1}^{L}\|\bar{w}_i\|_{l_1}$. This deduction can also be used to differ 2DRVFL from RVFL as well; (ii) Given similar level of sparsity in $\mathcal{W}$, SCN-based models are more prone to have a lower near-orthogonal level than RVFL-based ones, therefore,  indicating a smaller $\mbox{tr}(M)-\log\det(M)$.  For further justifications on these intuitive arguments, we present analogous theories regarding the near-orthogonality of weight vectors in the following part (see ST2 below) and demonstrate some statistical results at the end of this section.

\textbf{\emph{ST2: Weight Vector Angular Constraints for Diversity Promoting.}} Authors in \cite{Xie2015,Xie2017} have shown empirical effectiveness and explained in theory
when and why a low generalization error can be achieved via adjusting the diversity of hidden nodes in neural networks. Theoretically, increasing the diversity of hidden
nodes in a neural network model would reduce estimation error but increase approximation error, which implies that a low generalization error can be achieved when the diversity level is set appropriately. Specifically, near-orthogonality of the weight vectors (e.g, input weights and biases) can be used to characterize the diversity of hidden nodes in a neural network model, and a regularizer with weight vector angular constraints can be used to alleviate overfitting. To highlight the impact of near-orthogonality (of the weight vectors) on the generalization error,  we will reformulate two main theoretical results addressed in \cite{Xie2017}. Before that, some notations and preliminaries on statistical learning theory are revisited.

Consider the hypothesis set
\begin{eqnarray*}
  \mathcal{F} &:=\{x\mapsto \!\!&\sum_{j=1}^{L}\beta_jg(\bar{w}_j^{\mathrm{T}}x)\:\:\big|\:\: \|\beta\|_2\leq B, \|\bar{w}_j\|_2\leq C,\\
  &&\forall i\neq j, |\bar{w}_i^{\mathrm{T}}\bar{w}_j|\leq \tau\|\bar{w}_i\|_2\|\bar{w}_j\|_2\}.
\end{eqnarray*}
where $\beta$ stands for the output weight $g(t)=1/(1+e^{-t})$ is the sigmoid activation function.
Given training samples $\{(x_i,y_i)\}_{i=1}^N$ generated independently from an unknown distribution $\mathcal{P}_{\mathcal{X}\mathcal{Y}}$. The generalization error of $f\in \mathcal{F}$ is defined as $R(f)=\mathds{E}_{\mathcal{P}_{\mathcal{X}\mathcal{Y}}}[\frac{1}{2}(f(x)-y)^2]$. As $\mathcal{P}_{\mathcal{X}\mathcal{Y}}$ is not available, one can only consider minimizing the empirical risk $\hat{R}(f)=\frac{1}{2N}\sum_{i=1}^N(f(x_i)-y_i)^2$ in lieu of $R(f)$. Let $f^{*}\in \arg\min_{f\in \mathcal{F} }R(f)$ be the true risk minimizer and $\hat{f}\in \arg\min_{f\in \mathcal{F} }\hat{R}(f)$ be the empirical risk minimizer. Then, the generalization error $R(\hat{f}):=R(\hat{f})-R(f^{*})+R(f^{*})$ (of the empirical risk minimizer $\hat{f}$) can be estimated by bounding the estimation error $R(\hat{f})-R(f^{*})$ and the approximation error $R(f^{*})$, respectively. The following Theorem 2 and Theorem 3 show these two estimations in relation to the factor $\tau$.

\textbf{Theorem 2 \cite{Xie2017} (Estimation Error).} With probability at least $1-\delta$, the estimation upper bound of estimation error decreases as $\tau$ becomes smaller, i.e.,

\begin{equation*}
  R(\hat{f})-R(f^*)\leq\frac{\gamma^2\sqrt{2\ln(4/\delta)}+\gamma B(2C+4|g(0)|)\sqrt{m}}{\sqrt{N}},
\end{equation*}
where $\gamma=1+BC\sqrt{(m-1)\tau+1}/4+\sqrt{m}B|g(0)|$.

Suppose the target function $G=\mathds{E}[y|x]$ satisfy certain smoothness condition given by $\int\|\omega\|_2|\tilde{G}(\omega)|d\omega\leq B/2$, where $\tilde{G}(\omega)$ represents the Fourier transformation of $G$. Then, the approximation error, which reflects the power of the hypothesis set $\mathcal{F}$ for approximating $G$, is expressed as follows.

\textbf{Theorem 3 \cite{Xie2017} (Approximation Error).} A smaller $\tau$ contributes to a larger upper bound of approximation error, that is, let $C>1$ and $L\leq 2(\lfloor\frac{\pi/2-\theta}{\theta}\rfloor+1)$, where $\theta=\arccos (\pi)$, then there exists $f\in \mathcal{F}$ such that
\begin{equation*}
  \|f-G\|_2\!\leq\! B(\!\frac{1}{\sqrt{m}}+\frac{1+2\ln C}{C}\!)+2\sqrt{L}BC\sin(\frac{\min(2L\theta,\pi)}{2}).
\end{equation*}

Based on Theorem 2 and Theorem 3, we can come to a conclusion, that is, a larger upper bound of generalization error can be caused by the case when the weight vectors are highly near-orthogonal with each other ($\tau$ is extremely small) or the situation that $\tau$ is close or equal to 1 (e.g., there exist two weight vectors that are linearly dependent). Therefore, given two obtained (randomized) learner models with roughly the same training performance, the one equipped with hidden weight vectors of high near-orthogonality is likely to result in worse generalization. On the other hand, our previous work \cite{LiandWang2016} reveals a key pitfall of RVFL networks that all high-dimensional data-independent random features are nearly orthogonal to each other with probability one. Fortunately, the supervisory mechanism used in SCN framework imposes an implicit relationship between each weight vector and can effectively reduce the probability of near-orthogonality. With all these clues, we can roughly explain why the leaner models produced by SCN and 2DSCN are more prone to result in a better generalization performance than RVFL and 2DRVFL. It would be interesting to organize rigourous theoretical analysis and extensive empirical study on differing the SCN framework from RVFL networks from this point of view. Also, it is meaningful  to think about weight vector angular constraints in the development of SCNs, for the purpose of the enhancement of generalization. To avoid losing keynote for this work, we leave these useful explorations to our future research.

\textbf{\emph{ST3: Vague Relationship between 2DSCN and DropConnect framework.}}
To effectively alleviate over-fitting and improve the generalization performance, Dropout has been proposed for regularizing fully connected layers within neural networks by randomly setting a subset of activations to zero during training \cite{Hinton2012,Srivastava2014}. DropConnect proposed by Wan et al. \cite{Wan2013} is the extension of Dropout in which each connection, instead of each output unit, can be dropped with certain probability. Technically, DropConnect can be viewed as similar to Dropout because they both perform dynamic sparsity within the learner model during the training phase, however, differs in that the sparsity-based concerns are imposed on the hidden input weights, rather than on the output vectors of a layer. That means the fully connected layer with DropConnect becomes a sparsely connected layer in which the connections are chosen at random during the training stage. Importantly, as noted in \cite{Wan2013}, the mechanism employed in DropConnect is not equivalent to randomly assigning a sparse hidden input weights matrix (and remain fixed) during the training process, which indirectly invalidates the effectiveness of RVFL and 2DRVFL method even when they use sparse weights in the hidden layer.

Intuitively, our proposed 2DSCN could be thought as related to DropConnect, in terms of the following points:
\begin{itemize}
  \item supervisory mechanism used in 2DSCN aims at incrementally configuring the weight vectors until convergence to a universal approximator, which is equivalent to the training objective of DropConnect;
  \item once random weight vectors in 2DSCN have many small elements close to zero, their functionality is similar to the sparsity mechanism imposed in DropConnect on the hidden weights;
  \item On the basis of the above two clues, the incremental process performed in 2DSCN can be viewed as similar to proceeding dynamic sparsity within the learner model during the training phase as used in DropConnect.
\end{itemize}

\begin{figure}[htbp]
\centering
\includegraphics[width=0.49\textwidth]{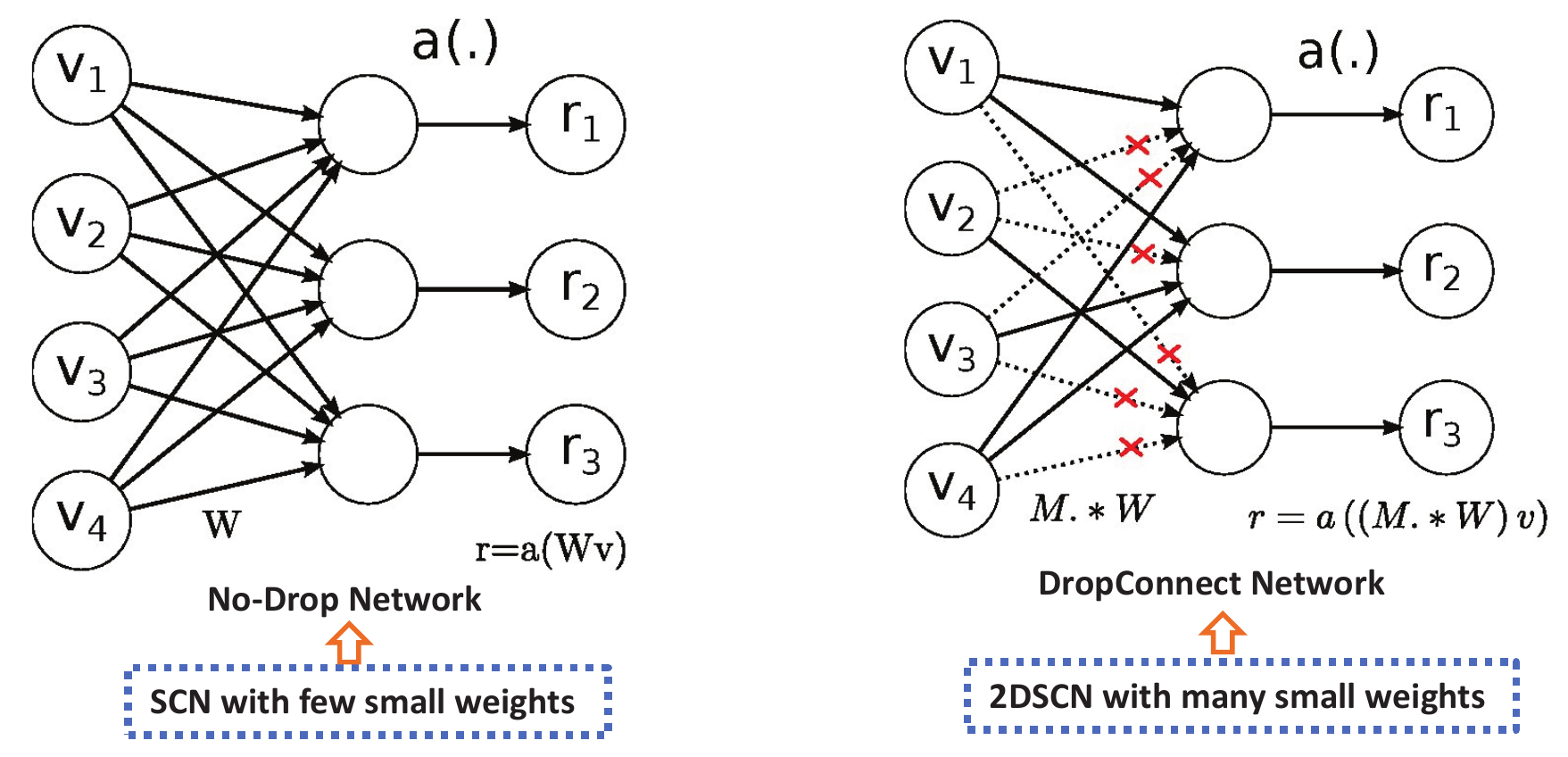}
\caption{Demonstration of No-Drop and DropConnect Network (screenshot from https://cs.nyu.edu/~wanli/dropc/) with our associations why 2DSCN differs from SCN (blue dotted boxes).}\label{DropConnect}
\end{figure}

We would like to highlight that the original SCN does not have this kind of vague relationship with DropConnect, unless certain weights sparsity regularizer is concerned in the training process. In contrast, 2DSCN involves more weight vectors with small values, which indeed can be viewed as considerable degree of sparsity, have a good potential to inherit some merits of DropConnect and its parallel methodology. Fig. \ref{DropConnect} highlights the characteristics of DropConnect, and provides a vivid demonstration of our logic why 2DSCN differs from SCN in exhibiting sparsity among the hidden input weights.

\textbf{\emph{ST4: Novel Estimation of Test Error.}}
Various statistical convergence rates for neural networks have been established when some constrains on the weights are concerned \cite{Bartlett1997,Bartlett1998, Bartlett2002,Zhong2017}. Empirically, small weights together with small training error can lead to significant improvements in generalization. All these investigations lend scientific supports to the heuristic techniques like weight decay and early stopping. The reason behind is that producing over-fitted mappings requires high curvature and hence large weights, while keeping the weights small during training can contribute to smooth mappings. Technically, the regularization learning framework, introducing various types of weight penalty such as L2 weight decay \cite{Hoerl1970,Poggio1985 }, Lasso \cite{Tibshirani1996}, Kullback-Leibler (KL)-divergence penalty \cite{Le2011}, etc., shares a similar philosophy to help prevent overfitting.

A comprehensive overview of existing theories/techniques concerning learner models's generalization capability is out of our focus in this paper. Instead, we revisit the theoretical result presented in \cite{Ali2009}, and illustrate mathematically how the output weights magnitudes affect randomized learner models' generalization power. For a better understanding and consistent problem formulation, we restate their main result with reference to our previous notations used in \emph{\textbf{ST2}}, that is,

\textbf{Theorem 4 \cite{Ali2009}.} Consider the hypothesis set $\mathcal{F}_p :=\{f(x)=\int\alpha(w)g(w;w)dw\big| |\alpha(w)|\leq Bp(w)\}$ with certain distribution $p$ and function $g$ satisfying  $\sup_{x,w}|g(x;w)|\leq 1$, and given a training data set with $N$ input-output pairs drawn iid from some distribution $\mathcal{P}_{\mathcal{X}\mathcal{Y}}$, a randomized learner model $\hat{f}(x)=\sum_{j=1}^{L}g(x_i;w_i)$ can be obtained by randomly assigning $w_i,w_2,\ldots,w_L$ from the distribution $p$ and solving the empirical risk minimization problem \footnote{*In [34], a general form of cost function is concerned. Here we specify a quadratic loss function and its associated Lipschitz constant has no impacts on the final estimation.} $\min_{\beta}\frac{1}{N}\sum_{i=1}^{N}(\hat{f}(x_i)-y_i)^2$ subject to $\|\beta\|_{\infty}\leq B/L$.
Then, with probability at least $1-2\delta$, the upper bound for the generalization error of $\hat{f}$ can be estimated by
\begin{equation*}
R[\hat{f}]\leq \min_{f\in \mathcal{F}_p}R[f]+\mathcal{O}\Bigg(\bigg(\frac{1}{\sqrt{N}}+\frac{1}{\sqrt{L}}\bigg)2B\sqrt{\log\frac{1}{\delta}}\Bigg)
\end{equation*}

Theoretically, the upper bound in Theorem 3 implies that randomized learner models with good training result and small output weights can probably lead to preferable generalization performance, in terms of probability perspective. However, this cannot be used directly to bound the practical test error for evaluating the randomized learner models's generalization performance. More numerical estimation for the test error (resulted from algorithm realization in practice) is required to better characterize the generalization capability as well as the associated impacting factors.

As one of our main contributions in this work, a novel upper bound estimation for the test error is presented in terms of computational perspective. To facilitate our theoretical investigation, we view the hidden layer matrix $H$ as a matrix-valued function of matrix variable, i.e., $H: \mathds{R}^{N\times d}\rightarrow \mathds{R}^{N\times L}$, denoted by (see \cite{Dattorro2010} for basic fundamentals on matrix calculus)
\begin{equation}\label{hiddenmatirx_1}
H:=H(X)=\left(
\begin{array}{ccc}
g(w^{\mathrm{T}}_1x_{1}+b_{1}) & \cdots & g(w^{\mathrm{T}}_Lx_{1}+b_{L}) \\
\vdots & \cdots & \vdots\\
g(w^{\mathrm{T}}_1x_{N}+b_{1}) & \cdots & g(w^{\mathrm{T}}_Lx_{N}+b_{L})
\end{array}\right)
\end{equation}
with the argument $X$ represented by
\begin{equation}\label{trainingsamplematrix}
X=(x_1,x_2,\ldots,x_N)^{\mathrm{T}}=\left(
\begin{array}{ccc}
x_{1,1} & \cdots & x_{1,d} \\
\vdots & \cdots & \vdots\\
x_{N,1} & \cdots & x_{N,d}
\end{array}\right)
\end{equation}
Suppose that $H$ is differentiable and has continuous first-order gradient $\nabla H$, defined by a quartix belongs to $\mathds{R}^{N\times L \times N \times d}$, i.e.,
\begin{equation*}
\nabla H(X)\!=\!\!\left(
\begin{array}{ccc}
 \nabla H_{1,1}(X)& \cdots &  \nabla H_{1,L}(X)  \\
\vdots & \cdots & \vdots\\
\nabla H_{N,1}(X)& \cdots &  \nabla H_{N,L}(X)
\end{array}\right),
\end{equation*}
where for $i=1,2,\ldots,N$, $j=1,2,\ldots,L$
\begin{equation*}
\nabla H_{i,j}(X)=\left(
\begin{array}{ccc}
 \frac{\partial g(w^{\mathrm{T}}_jx_{i}+b_{j}) }{\partial x_{1,1}}& \cdots & \frac{\partial g(w^{\mathrm{T}}_jx_{i}+b_{j})}{\partial x_{1,d}}   \\
\vdots & \cdots & \vdots\\
\frac{\partial g(w^{\mathrm{T}}_jx_{i}+b_{j})}{\partial x_{N,1}}& \cdots & \frac{\partial g(w^{\mathrm{T}}_jx_{i}+b_{j})}{\partial x_{N,d}}
\end{array}\right).
\end{equation*}
Then, the first directional derivative in a given direction $Z\in \mathds{R}^{N\times d}$ can be represented by
\begin{eqnarray*}
&&\overset{\rightarrow Z}d\!\!H(X):=\\&&\left(
\begin{array}{ccc}
\mbox{tr}\Big(\nabla H_{1,1}(X)^\mathrm{T}Z\Big) & \cdots & \mbox{tr}\Big(\nabla H_{1,L}(X)^\mathrm{T}Z \Big) \\
\vdots & \cdots & \vdots\\
\mbox{tr}\Big(\nabla H_{N,1}(X)^\mathrm{T}Z\Big) & \cdots & \mbox{tr}\Big(\nabla H_{N,L}(X)^\mathrm{T}Z \Big)
\end{array}\right)
\end{eqnarray*}
It is logical to think that the test sample matrix $\tilde{X} $ can be represented by imposing sufficiently small random noises into the training sample matrix $X$, i.e., $\tilde{X}:=X+\epsilon Z$, where $Z\in \mathds{R}^{N\times L}$ is a random matrix, $\epsilon$ is sufficiently small.
Then, we can take the first-order Taylor series expansion about $X$ (\cite{Dattorro2010}), i.e.,
\begin{equation*}
  H(\tilde{X}):=H(X+\epsilon Z)=H(X)+\epsilon\!\!\!\!\!\overset{\:\:\:\:\:\rightarrow Z}d\!\!H(X)+o(\epsilon^2)
\end{equation*}
Therefore, the test error can be estimated by
\begin{eqnarray}\label{test_error}
  &&\|H(\tilde{X})\beta-Y\|_{F}\nonumber\\
  &=& \|(H(X)+\epsilon\!\!\!\!\!\overset{\:\:\:\:\:\rightarrow Z}d\!\!H(X)+o(\epsilon^2))\beta-Y\|_{F}\nonumber \\
  &\leq& \|H(X)\beta-Y\|_{F}+\epsilon\|\!\!\!\!\overset{\:\:\:\rightarrow Z}d\!\!H(X)\|_{F}\|\beta\|_{F}\nonumber\\&&+o(\epsilon^2)\|\beta\|_{F}
\end{eqnarray}
where $\|\cdot\|_{F}$ stands for the Frobenius norm, $ \|H(X)\beta-Y\|_{F}$ represents the training error. \newline
Basically, this rough estimation implies two points that should be highlighted:\\
(i) The upper bound for the test error can be viewed as an increasing function of $\|\beta\|_{F}$, which means that learner models with smaller output weight values are more prone to generalize preferably on unseen data. This is consistent with the philosophy behind the regularization learning framework, that is, imposing a penalty term to control the output weights magnitudes during the training process.\\
(ii) We can further investigate how the input weights and biases affect the value of $\|\!\!\!\!\overset{\:\:\:\rightarrow Z}d\!\!H(X)\|_{F}$. In particular, we use sigmoid function in the following deduction, i.e., $g(t)=1/(1+e^{-t})$ and $g'(t)=g(t)(1-g(t))$. Mathematically, the $(i,j)$-th element ($i=1,2,\ldots,N$, $j=1,2,\ldots,L$) inside \!\!\!\!$\overset{\:\:\:\:\rightarrow Z}d\!\!H(X)$ can be expressed as
\begin{eqnarray*}
 &&\mbox{tr}\Big(\nabla H_{i,j}(X)^\mathrm{T}Z\Big)\\&:=&\sum_{i^{'}=1}^{N}\sum_{k^{'}=1}^{d}\frac{\partial g(w^{\mathrm{T}}_jx_{i}+b_{j})}{\partial x_{i^{'},k^{'}}}Z_{i^{'},k^{'}}\\
 &=&-g(w^{\mathrm{T}}_jx_{i}+b_{j})(1-g(w^{\mathrm{T}}_jx_{i}+b_{j}))\sum_{k=1}^{d}w_{j,k}Z_{i,k}.
\end{eqnarray*}
Then, a rough upper bound for  $\|\!\!\!\!\overset{\:\:\:\rightarrow Z}d\!\!H(X)\|_{F}$ can be obtained, that is,
\begin{eqnarray*}\label{bound}
 &&\|\!\!\!\!\overset{\:\:\:\rightarrow Z}d\!\!H(X)\|_{F}\nonumber\\&:=&\sqrt{\sum_{i=1}^{N}\sum_{j=1}^{L} \mbox{tr}\Big(\nabla H_{i,j}(X)^\mathrm{T}Z\Big)^2}\nonumber\\
 &=&\sqrt{\sum_{i=1}^{N}\sum_{j=1}^{L} (g_{ij}(1-g_{ij}))^2(\sum_{k=1}^{d}w_{j,k}Z_{i,k})^2}\nonumber\\
 &\leq&\sqrt{\sum_{i=1}^{N}\sum_{j=1}^{L}(g_{ij}(1-g_{ij}))^2(\sum_{k=1}^{d}w_{j,k}^2)(\sum_{k=1}^{d}Z_{i,k}^2)}\nonumber\\
 &=&\sqrt{\sum_{i=1}^{N}\sum_{j=1}^{L}(g_{ij}(1-g_{ij}))^2\|w_j\|_2^2\|Z_i\|_2^2}\nonumber\\
 &\leq & \max_{1\leq i \leq N}\|Z_i\|_2\cdot\|H\circ(O-H)\circ\ddot{W}\|_{F},
 \end{eqnarray*}
where we use abbreviation $g_{ij}$ for $g(w^{\mathrm{T}}_jx_{i}+b_{j})$), Cauchy-Schwarz inequality in the first inequality. $Z_i$ stands for the $i$-th row vector of the matrix $Z$. $H$ is defined in (\ref{hiddenmatirx_1}), $O\in \mathds{R}^{N\times L}$ is a matrix of ones (every element is equal to one), and $\ddot{W} \in \mathds{R}^{N\times L}$ is formulated by copying $N$-times of the row vector $(\|w_{1}\|_2, \|w_{2}\|_2,\ldots,\|w_{L}\|_2)$, `$\circ$' stands for the Hadamard (entrywise) product among the matrixes.

So far, we can summarize the above theoretical result in the following Theorem 5. Readers can refer to some notations aforementioned in the context.

\textbf{Theorem 5.} Given training input $X\in \mathds{R}^{N\times d}$ and output $Y\in \mathds{R}^{N\times m}$, suppose a randomized neural network model with $L$ hidden nodes is build, corresponding to the hidden layer output matrix (on the training data) $H\in \mathds{R}^{N\times L}$, the output weight matrix $\beta$, and the training error $\|H\beta-Y\|_{F}$. Let $\tilde{X}:=X+\epsilon Z$ be the test (unseen) input data matrix, where $Z\in \mathds{R}^{N\times L}$ is a random matrix, $\epsilon$ is sufficiently small, $\tilde{H}$ stand for the associated hidden layer output matrix, then, the test error can be bounded by
\begin{eqnarray}\label{theorem4}
 &&\|\tilde{H}\beta\!-\!\!Y\|_{F}\nonumber\\
 &\leq&\|H\beta\!-\!\!Y\|_{F}+\epsilon\!\!\max_{1\leq i \leq N}\!\!\|Z_i\|_2\!\cdot\!\|H\!\circ\!(O\!-\!H)\!\circ\!\ddot{W}\|_{F}\|\beta\|_{F}\nonumber\\&&+o(\epsilon^2)\|\beta\|_{F}.
\end{eqnarray}

\textbf{Remark 3.} We would like to highlight a trick concerned in the previous deduction for Theorem 5. Indeed, we have considered to preserve the bundle of computational units `$(g_{ij}(1-g_{ij}))^2\|w_j\|_2^2$' rather than to roughly estimate the whole term by `$(\frac{1}{4})^2\|w_j\|_2^2$', which consequently can result in a very blunt bound $\frac{1}{4}\|W\|_{F}\|Z\|_{F}$ for $\|\!\!\!\!\overset{\:\:\:\rightarrow Z}d\!\!H(X)\|_{F}$. Unfortunately, upper bound $\frac{1}{4}\|W\|_{F}\|Z\|_{F}$ sounds meaningless because it does not consider the saturation property of sigmoid function, and may cause some misleading that `larger input weights can destroy the generalization capability'. In contrast, our proposed upper bound (\ref{bound}) is nearly sharp and can provide valuable information to identify the role of input weights (and biases) and training samples on the learner models's generalization power. It is the bundle of computational units $\|H\circ(O-H)\circ\ddot{W}\|_{F}$ rather than merely the $\|W\|_{F}$ that acts as a suitable indicator for predicting the generalization performance. Besides, it should be noted that, input weights (and biases) with small values but enforcing the $g(\cdot)\approx1$ or $g(\cdot)\approx0$ (corresponding to the saturation range of sigmoid function), are more likely to result in a small value of $\|H\circ(O-H)\circ\ddot{W}\|_{F}$ and consequently bring a small generalization error bound.

On the other hand, the right side of Eq. (\ref{theorem4}) has a strong resemblance to the regularized learning target by viewing $\epsilon\max_{1\leq i \leq N}\!\!\|Z_i\|_2\!\cdot\!\|H\!\circ\!(O\!-\!H)\!\circ\!\ddot{W}\|_{F}$ as the regularization factor $\sigma>0$, that is, $\|H\beta\!-\!\!Y\|_{F}+\sigma\|\beta\|_{F}$, considered as a whole to effectively alleviate over-fitting.

\textbf{\emph{Why 2D randomized models are equipped with more small weights?}} Since small weights to some extent can probably have certain positive influence on enhancing a learner model's generalization ability, one major issue still left unclarified is that whether or not 2D randomized learner models possess this advantage. For that purpose and before ending this section, we would like to provide a statistical verification on the frequency when sufficiently small weights occur in 1D and 2D randomized models, aiming to further support the superiority of 2D randomized models.

Given distribution $\mathcal{P}$ (either uniform or gaussian), we investigate the statistical differences among the following three strategies for randomly assign parameters:
\begin{itemize}
  \item M1: Randomly assign $w=[w_1,w_2,\ldots,w_d]^\mathrm{T}$ from $\mathcal{P}$;
  \item M2: Randomly assign $z_1=[z_{1,1},z_{1,2},\ldots,z_{1,d}]^\mathrm{T}$, $z_2=[z_{2,1},z_{2,2},\ldots,z_{2,d}]^\mathrm{T}$ from $\mathcal{P}$, then calculate their Hadamard (entrywise) product $w^{(1D-P)}=z_1\circ z_2$;
  \item M3: Randomly assign $u=[u_1,u_2,\ldots,u_{d_1}]^\mathrm{T}$, $v=[v_1,v_2,\ldots,v_{d_2}]^\mathrm{T}$, with $d_1d_2=d$, then calculate $uv^{\mathrm{T}}$ and let $w^{2D}:= vec(uv^{\mathrm{T}})$.
\end{itemize}

\begin{figure}[htbp!]
\centering
\subfigure[$w^{2D}_i$]{\includegraphics[width=0.24\textwidth]{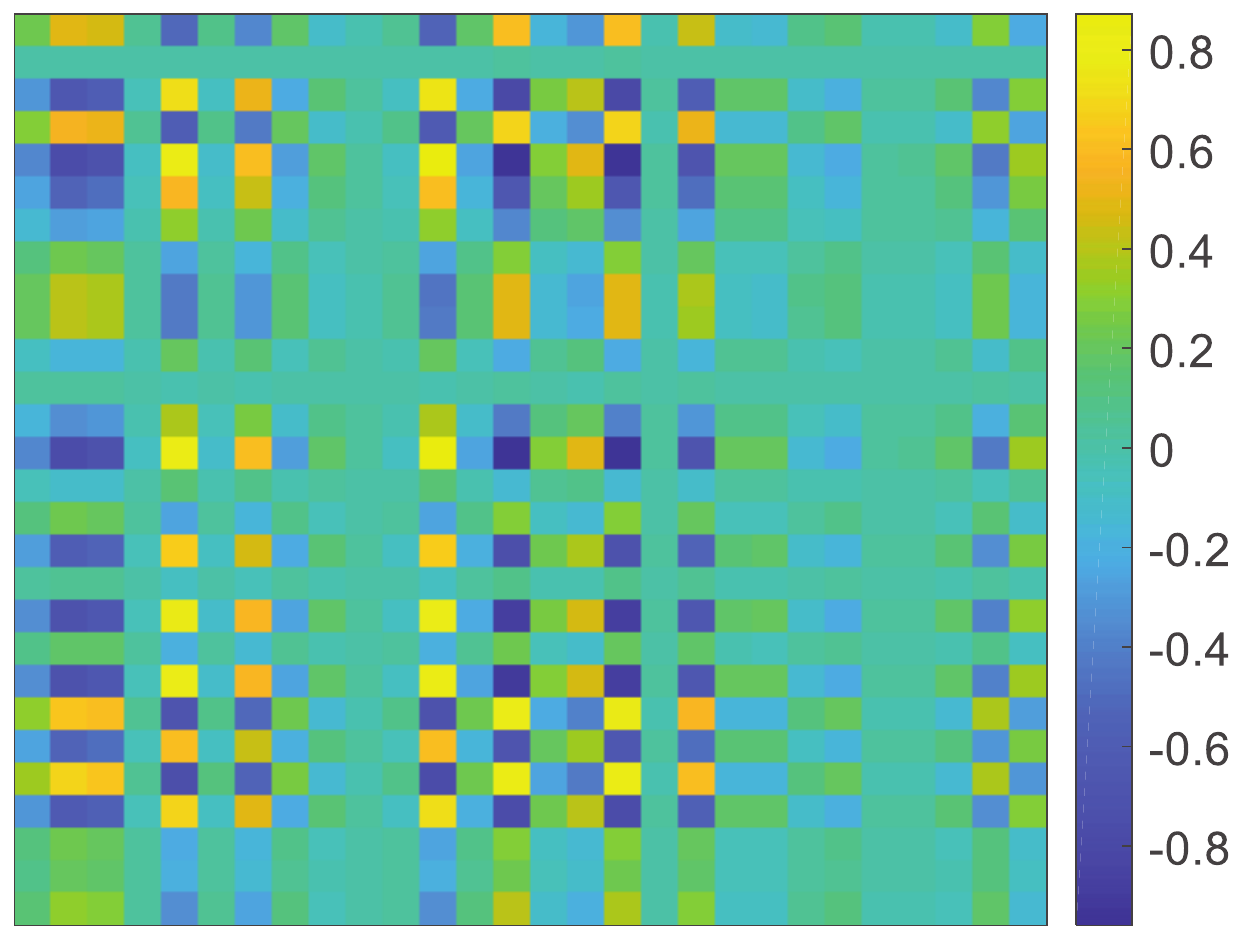}}
\subfigure[$w_i$]{\includegraphics[width=0.24\textwidth]{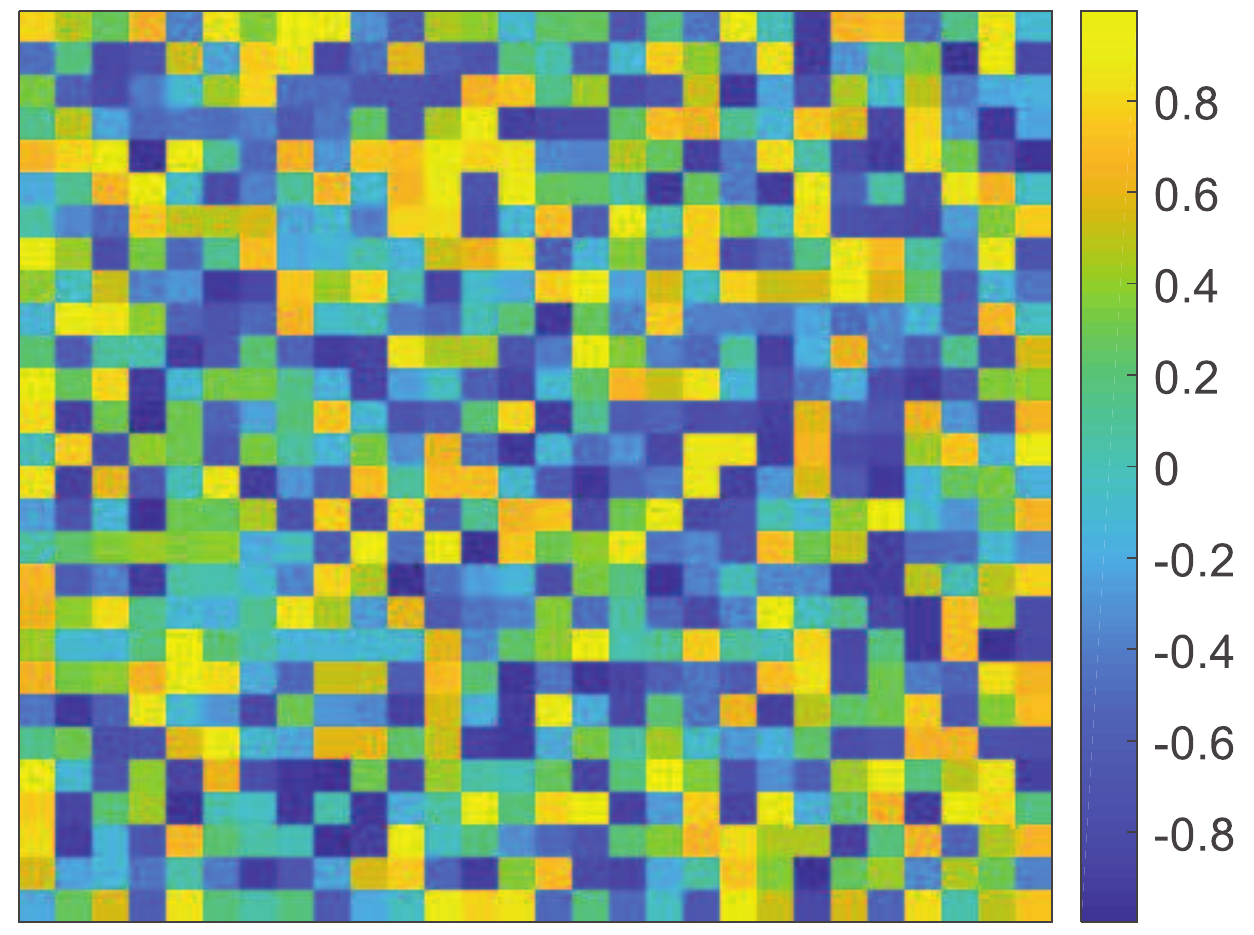}}
\caption{Simple illustration for the distribution of 1D and 2D random weights: (a) Values of $w^{2D}_i$ generated by M3 with $d_1=d_2=28$; (b) Values of $w_i$ generated by M1 with $d=784$. Both $w^{2D}_i$ and $w_i$ are reshaped into $28\times 28$ for visualization.}\label{Distribution2D-1D}
\end{figure}

A simple and vivid demonstration for the distribution of the random weights induced by M1 and M3 is provided in Fig. \ref{Distribution2D-1D}, in which it can be clearly seen that
$w^{2D}$ have more small values (near zero) than $w$. Based on our empirical experience, similar plotting (display) between  M2 and M3 looks visually indistinguishable. More statistical results are helpful for making a reasonable distinction among M1$\sim$M3.

Our primary objective is to find out how frequently can these strategies contribute to a high dimensional random vector ($w$, $w^{(1D-P)}$, or $w^{2D}$) with considerable amount of elements whose values are close to zero. Here we present a theoretical result for answering this question and then provide an empirical verification in statistics. Specifically, from a probability perspective, we conclude that \textbf{M3} are more prone to get a random vector with more elements close to 0, as mathematically expressed by
\begin{eqnarray}\label{propability}
  &&P\left(\frac{\#\Big\{i\Big||w^{2D}_i|\leq \epsilon\Big\}}{d}\geq p\right)\nonumber\\
  &\geq& P\left(\frac{\#\Big\{i\Big||w^{(1D-P)}_i|\leq \epsilon\Big\}}{d}\geq p\right)\nonumber\\
  &\geq& P\left(\frac{\#\Big\{i\Big||w_i|\leq \epsilon\Big\}}{d}\geq p\right),
\end{eqnarray}
where $\epsilon$ is a small value close to 0, as a reference factor for locating small elements of the random vector. $\#\{\cdot\}$ stands for the cardinality, which of course is equal to the number of elements we are interested in counting, i.e, whose absolute values equal/lower than $\epsilon$. On the other hand, $p$ is a threshold by which we can study the normalized percentage ($\#\{\cdot\}/d\in [0,1]$) when interested elements occur in that random vector ($w$, $w^{(1D-P)}$, or $w^{2D}$).

Instead of making efforts to give a rigourous mathematical proof, here we focus on a empirical study to verify the inequalities (\ref{propability}) in statistics.  In particular, we set $d_1=d_2=28$, $d=784$, $p=8\%, 10\%, 12\%, 15\%$, $\epsilon=0.01, 0.03, 0.05, 0.1$, respectively, and study two options for $\mathcal{P}$, i.e., uniform (Case 1) and gaussian distribution (Case 2), then run 100,000 independent numerical simulations to approximate those three probability values compared in (\ref{propability}), in terms of each set of $(p,\epsilon)$ for both distribution cases. In the following, we give some theoretical description and the statistical results for Case 1 and 2.

\emph{Case 1}. Given two independent uniform random variables $z_1\sim U[-1,1]$ and $z_1\sim U[-1,1]$,  the probability density function (p.d.f) of their product $z=z_1z_2$ can be expressed by
\begin{equation*}
  p(x)=\begin{cases}
\frac{-1}{2}\ln x, & 0 < z\leq 1\\
\frac{-1}{2}\ln (-x), & -1\leq z < 0
\end{cases}
\end{equation*}
In the simulation, we conduct 100,000 independent trials for randomly assigning $w$, $w^{(1D-P)}$, and $w^{2D}$ according to M1$~$M3, respectively. Later we can count the number of times (denoted by $M$) when the condition $\#\{\cdot\}/d\geq p$ are satisfied, followed by roughly estimating the true probability in (\ref{propability}) with $\tilde{P}=M/100000$. In Table \ref{statistical_1}, we list the corresponding $\tilde{P}$ values (arranged in order $w^{2D}/w^{(1D-P)}/w$) for the cases with different settings of $\epsilon$ (e.g., $0.001, 0.005, 0.01$) and $p$ (e.g., $8\%, 10\%, 12\%, 15\%$), demonstrating that 2D models have more opportunities to have small input weights during the training process.
\begin{table}[htbp!]
\renewcommand{\arraystretch}{1.35}
	\caption{Statistical Verification for Case 1 Uniform Distribution}\label{statistical_1}
	\centering
	\begin{tabular}{|c|c|c|c|}
		\hline
		Case 1 & $\epsilon=0.001$  & $\epsilon=0.005$ & $\epsilon=0.01$ \\
		
		\hline
		$p=8\%$ &	0.0047\:/\:0\:/\:0  &0.1338\:/\:2.0e-5\:/\:0&0.4022\:/\:0.2838\:/\:0 \\
		\hline
		$p=10\%$ &	1.1e-4\:/\:0\:/\:0  &0.0160\:/\:0\:/\:0&0.1131\:/\:0\:/\:0 \\
		\hline
        $p=12\%$ &	3.0e-5\:/\:0\:/\:0  &0.0045\:/\:0\:/\:0&0.0497\:/\:0\:/\:0 \\
		\hline
		$p=15\%$ &	0\:/\:0\:/\:0  &6.7e-4\:/\:0\:/\:0&0.0126\:/\:0\:/\:0 \\
		\hline
	\end{tabular}
\end{table}

\emph{Case 2}. Given two independent normal random variables $z_1\sim N(0,\sigma_1^2)$ and $z_2\sim N(0,\sigma_2^2)$, the probability density function (p.d.f) of their product $z=z_1z_2$ can be expressed by
\begin{equation*}
  p(x):=\frac{1}{\pi\sigma_1\sigma_2}K_0\left(\frac{|x|}{\sigma_1\sigma_2}\right), x\in \mathds{R},
\end{equation*}
where $K_0(\cdot)$ is a modified Bessel function of the second kind of order zero \cite{Springer1970}, as given by
\begin{equation*}
  K_0(x)=\int_{0}^{\infty}e^{-x\cosh(t)}dt
\end{equation*}
Similarly, we present the associated $\tilde{P}$ values for $w^{2D}/w^{(1D-P)}/w$ respectively in Table \ref{statistical_2}, in which the records also show that 2D randomized models are more prone to be equipped with small weights.
\begin{table}[htbp!]
\renewcommand{\arraystretch}{1.35}
	\caption{Statistical Verification for Case 2 Gaussian Distribution}
	\label{statistical_2}
	\centering
	\begin{tabular}{|c|c|c|c|}
		\hline
		Case 2 & $\epsilon=0.001$ & $\epsilon=0.005$ & $\epsilon=0.01$  \\		
		\hline
		$p=8\%$ &	0.0011\:/\:0\:/\:0  &0.0445\:/\:0\:/\:0&0.1781\:/\:4.6e-4\:/\:0 \\
		\hline
		$p=10\%$ &	1.0e-5\:/\:0\:/\:0  &0.0025\:/\:0\:/\:0&0.0248\:/\:0\:/\:0 \\
		\hline
        $p=12\%$ &	0\:/\:0\:/\:0  &3.8e-4\:/\:0\:/\:0&0.0071\:/\:0\:/\:0 \\
		\hline
		$p=15\%$ &	0\:/\:0\:/\:0  &2.0e-5\:/\:0\:/\:0&9.2e-4\:/\:0\:/\:0 \\
		\hline
	\end{tabular}
\end{table}
\section{Performance Evaluation}
In this section, we demonstrate the advantages of 2DSCNs in image data modelling tasks, compared with some baseline/randomized learning methods including SCN, RVFL and 2DRVFL networks. Both regression and classification problems with image inputs are examined in the simulation. Datasets description, experimental setup, results and discussions are detailed for each task, as introduced in the following.
\subsection{Regression: Rotation Angles Predication for Handwritten Digits}
We first demonstrate the merits of the proposed 2DSCN by predicting the angles of rotation of handwritten digits. In particular, Neural Network Toolbox in MATLAB R2017b provides a collection of synthetic handwritten digits, which contains 5000 training and 5000 test images of digits with corresponding angles of rotation. Each image represents a rotated digit in grayscale and of normalized size ($28 \times28$). For example, 16 random samples are displayed in Fig. \ref{demo}. Apparently, baseline approaches such as RVFL and SCN deploy one dimensional input (reshaping the image into a vector) in modelling, while 2DRVFL and 2DSCN can directly deal with image-based inputs in problem-solving. It should be noted that, $w$ in RVFL, $u$ (and $v$) in 2DRVFL are randomly assigned from $[-1,1]^{784}$ and $[-1,1]^{28}$, respectively, while the biases for them are randomly assigned from $[-1,1]$. For SCN and 2DSCN, we take $T_{max}=5$, $\lambda=\{1,5,15,30,50,100,150,200,250\}$, $r=\{1-10^{-j}\}_{j=2}^{7}$ in the algorithm implementation (see Section 3 for the functionality of these parameters).
\begin{figure}[htbp]
\centering
\includegraphics[width=0.3\textwidth]{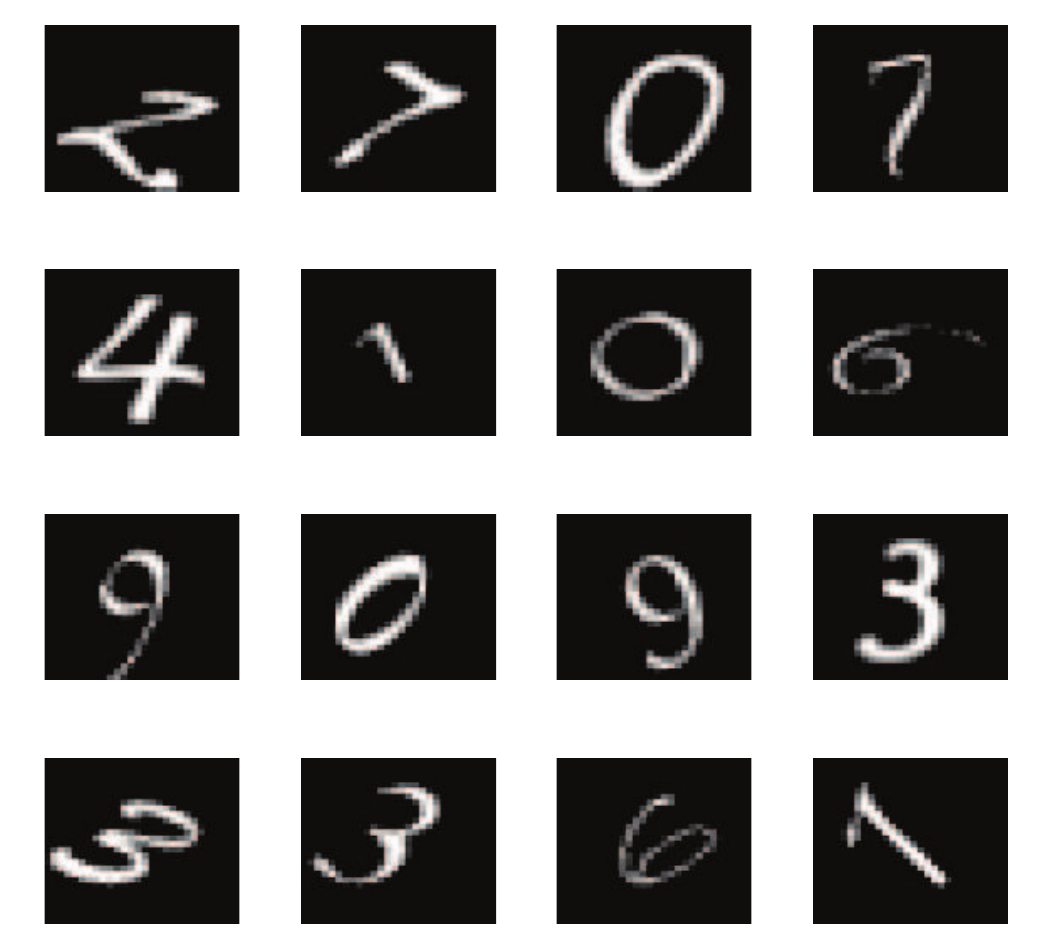}
\caption{Samples of Handwritten Digits with Angles of Rotation.}\label{demo}
\end{figure}

Two types of evaluation metrics are used in performance comparison: (i) The Percentage of Predictions within an Acceptable error margin (PPA); and (ii) The root-mean-square error (RMSE) of the predicted and actual angles of rotation. In particular, a user-defined threshold $\theta$ (in degrees) is needed to measure PPA values, that is, calculating the error between the predicted and actual angles of rotation and then counting the number of predictions within an acceptable error margin $\theta$ from the true angles. Mathematically, the PPA value within threshold $\theta$ can be obtained by
\begin{equation*}
  PPA=\frac{\#\{|Prediction\:\:Error| < \theta\}}{Number\:\:of\:\:Sample\:\:Images}.
\end{equation*}
\begin{table}[htbp!]
\renewcommand{\arraystretch}{1.35}
	\caption{Performance Comparison on Training (Tr) and Test (Te)}
	\label{Regression_table1}
	\centering
	\begin{tabular}{|c|c|c|c|c|c|c|}
		\hline
		\multirow{2}{*}{Algorithms} & \multicolumn{2}{c|}{PPA ($\%$), $\theta=15$} & \multicolumn{2}{c|}{PPA ($\%$), $\theta=25$}& \multicolumn{2}{c|}{RMSE} \\
		\cline{2-7}
		& Tr & Te & Tr & Te & Tr & Te \\
		\hline
		RVFL &	95.15 & 79.97 &99.76 & 95.64 &7.51 & 12.04  \\
		\hline
		SCN &99.84 &87.19 &99.98&  98.08 &4.57 & 10.08\\
		\hline
        2DRVFL &95.59	& 81.05&99.78& 95.72 &7.32 & 11.91  \\
		\hline
		2DSCN  &\textbf{99.88}  & \textbf{87.55} & \textbf{100} & \textbf{98.12 } &\textbf{4.42} & \textbf{9.96}\\
		\hline
	\end{tabular}
\end{table}
Table \ref{Regression_table1} shows both training and test results for these four algorithms ($L=1800$) in terms of PPA and RMSE, with threshold $\theta$ set to 15 and 20, respectively. It is observed that both SCN and 2DSCN outperform RVFL-based algorithms, and 2DSCN shows the highest PPA and lowest RMSE values, which reflects a better learning and generalization capability. Specifically, given $\theta=25$, 2DSCN-based learner model contributes $100\%$ (training) and  $98.12\%$ (test) PPA values, compared with 2DRVFL with $99.78\%$ and $95.72\%$, respectively. Based on simple calculation according to PPA values and the number of training/test samples, we can immediately notice that 2DRVFL produces some training errors with absolute values larger than 25 degrees while all the training predictive errors of 2DSCN are under this threshold. Also, there are less than 100 test digits predicted by 2DSCN, however, more than 200 instances led by 2DRVFL, with degree errors outside the interval $[-25,25]$. It should be noted that all the results illustrated in Table \ref{Regression_table1} are the averaged values based on 50 independent trials. Based on a rule of thumb, their standard deviations have no significant difference, so we omit this information here without any confusion.
\begin{figure}[htbp]
\centering
\subfigure[Training]{\includegraphics[width=0.24\textwidth]{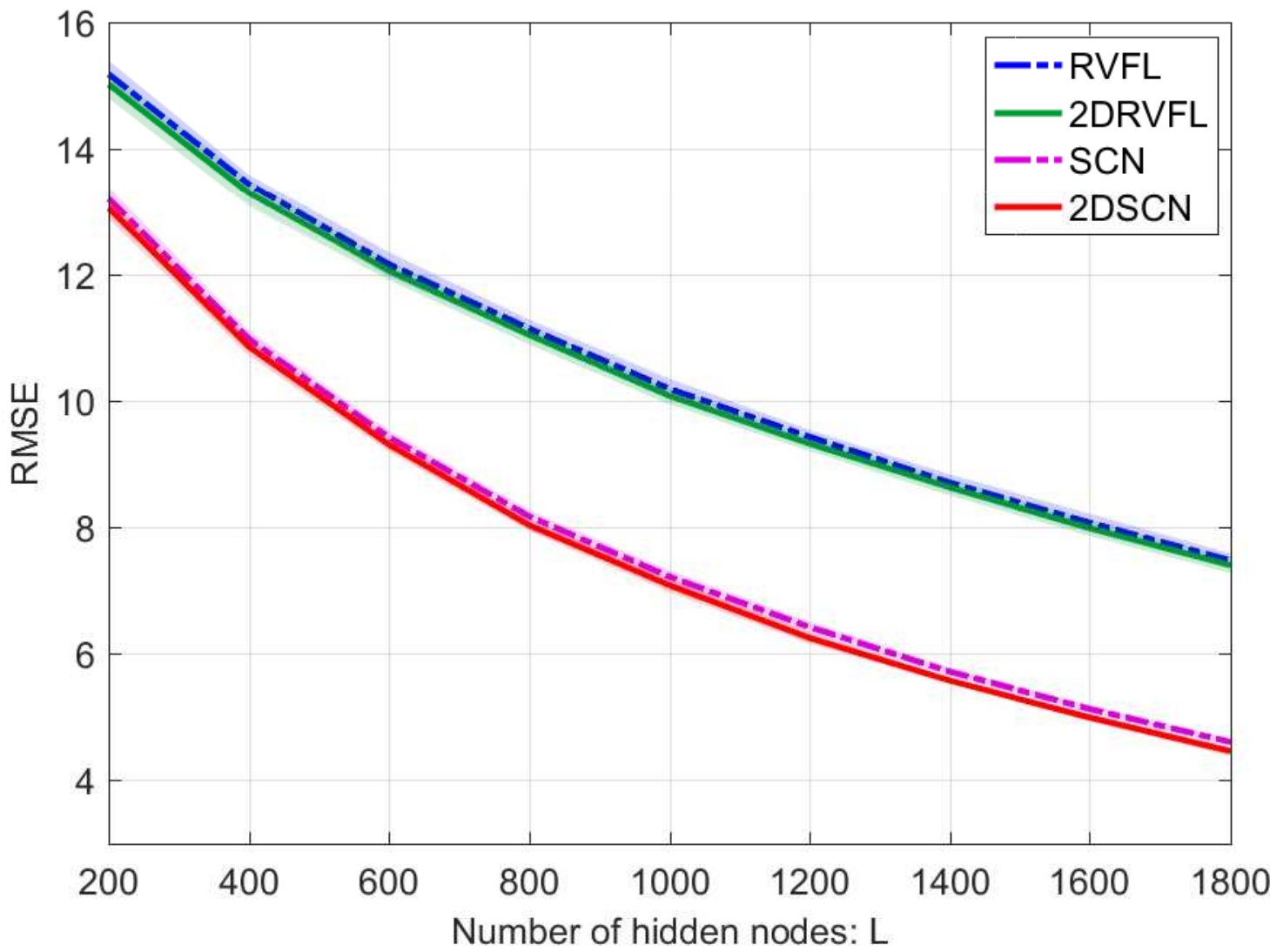}}
\subfigure[Test]{\includegraphics[width=0.24\textwidth]{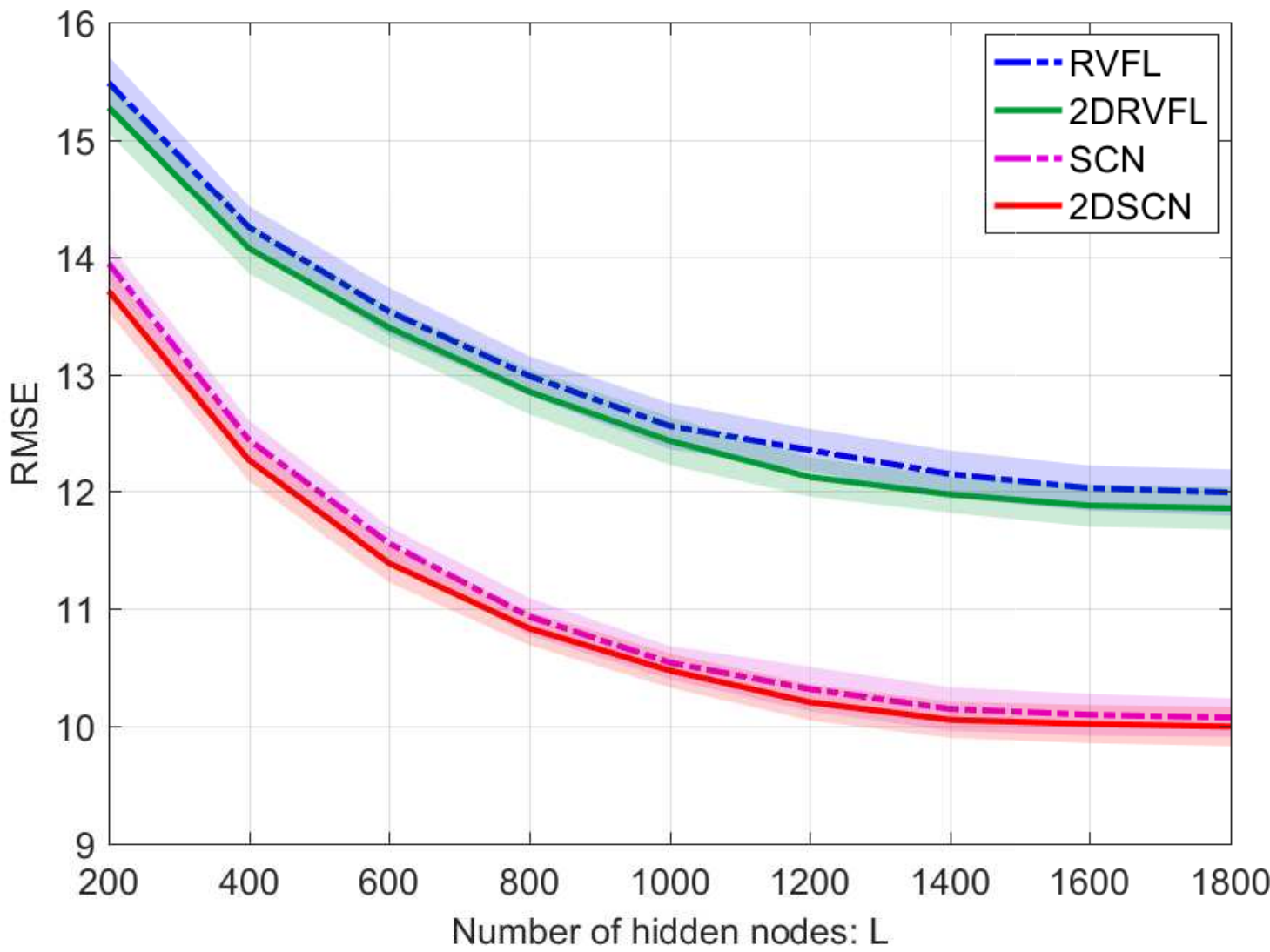}}
\caption{Performance comparison with different setting of $L$ on both training and test}
\end{figure}
\begin{figure}[htbp]
\centering
\subfigure[2DRVFL]{\includegraphics[width=0.24\textwidth]{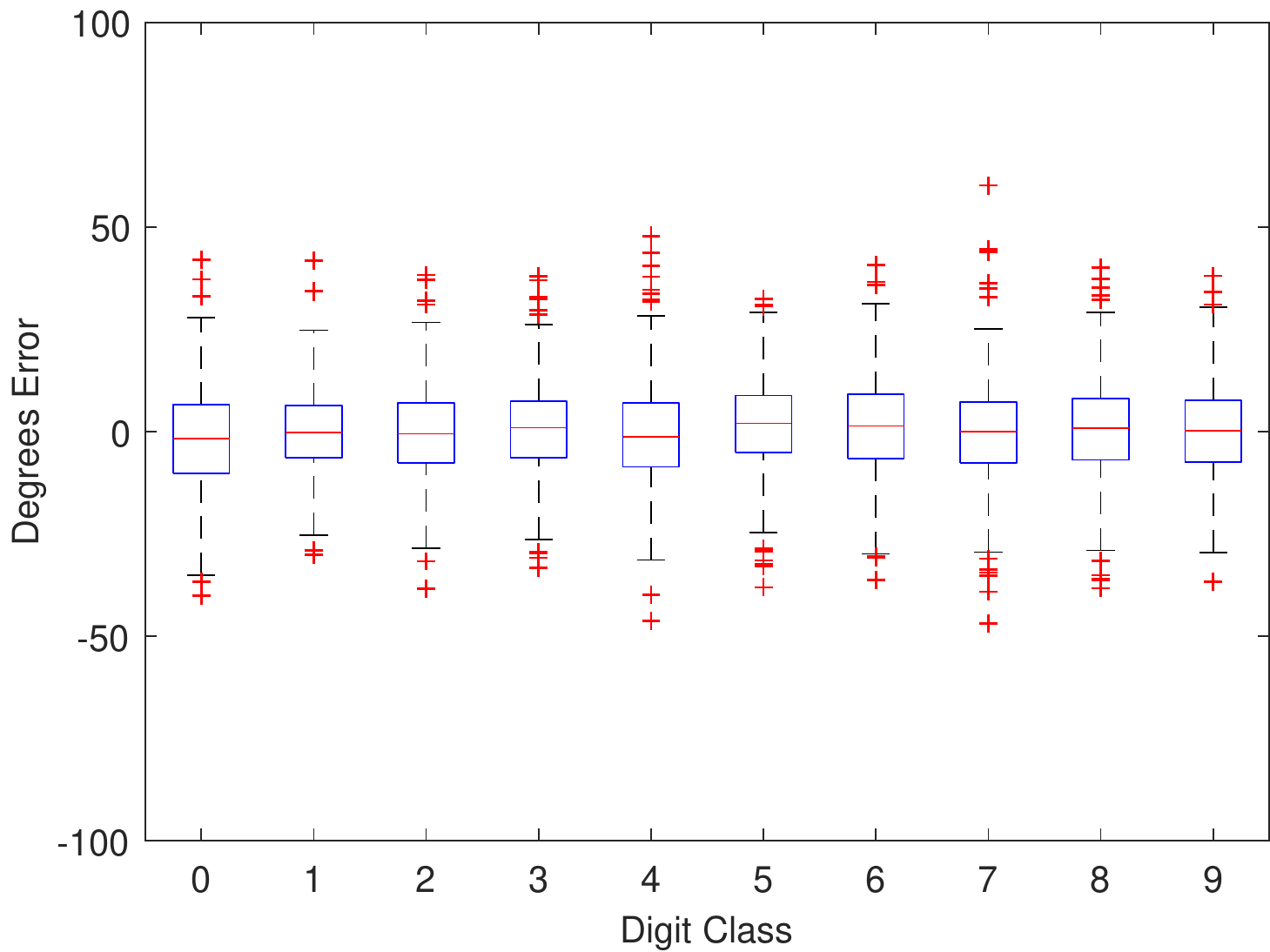}}
\subfigure[2DSCN]{\includegraphics[width=0.24\textwidth]{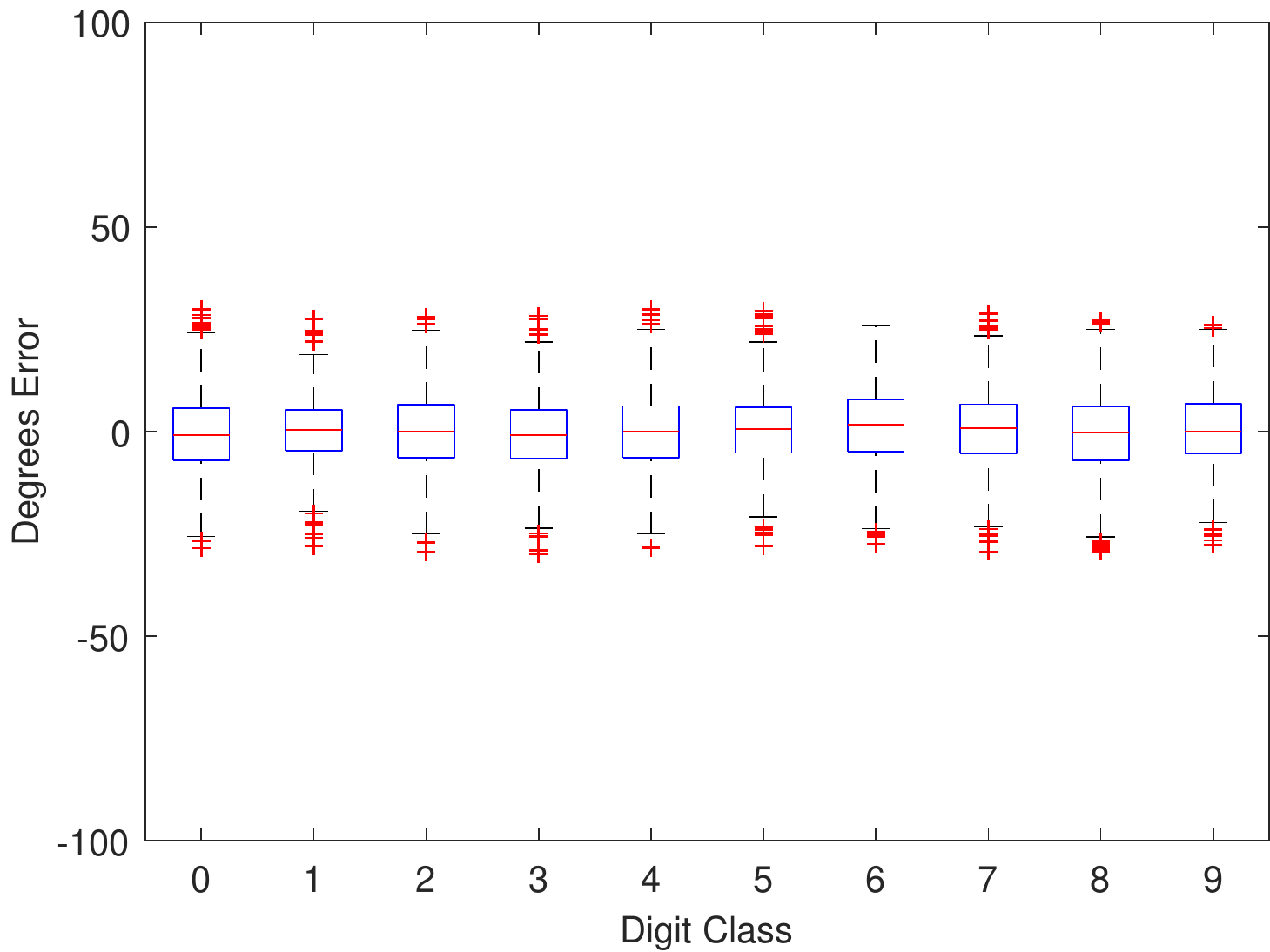}}
\caption{Predictive error comparison between 2DSCN and 2DRVFL for each digit}
\end{figure}
To fully demonstrate the difference between 2DSCN and 2DRVFL on residual distribution for test images (for each digit class), we plot the corresponding box-and-whisker diagram in Fig. 6. It is obvious that 2DSCN works much more favorably than 2DRVFL for almost each digit class as there are less records with abnormal degree error marked in red plus symbol. 2DSCN shows better stability on prediction for every digit as there are less errors outside the interval $[-40,40]$ than that of 2DRVFL. One can easily observe that this finding also corresponds what we have presented in Table \ref{Regression_table1}. In particular, the learner model produced by 2DRVFL is more prone to result in larger predictive errors (with absolute values close/higher than 50) for digit '4' and '7'. Obviously, 2DSCN can contribute to a slightly better rotation correction than SCN for each setting of the learner model architecture, while they both outperform RVFL and 2DRVFL.

\subsection{Classification: Handwritten Digits Recognition}
\begin{figure*}[htbp!]
\centering
\subfigure[Bangla]{\includegraphics[width=0.46\textwidth]{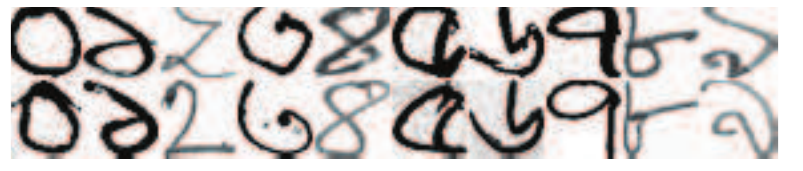}}\hspace{6mm}
\subfigure[Devnagari]{\includegraphics[width=0.46\textwidth]{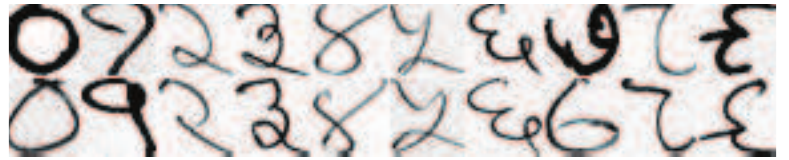}}\hspace{6mm}
\subfigure[Oriya]{\includegraphics[width=0.46\textwidth]{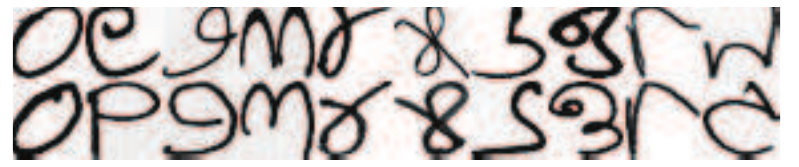}}\hspace{6mm}
\subfigure[CVL]{\includegraphics[width=0.46\textwidth]{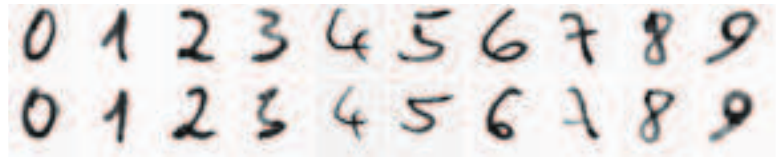}}
\caption{Samples of handwritten digits from Bangla, Devnagari, Oriya, and CVL database.}\label{fig:handwritten}
\end{figure*}

In this part, we compare our proposed 2DSCN algorithm with the other three randomized approaches on image classification problem. Four benchmark handwritten digits databases are employed in the comparison. Parameter setting for these four algorithms ($w,b,u,v,\lambda,T_{max},r$) are the same as the configuration used in the previous regression task. In particular, Bangla, Devnagari, and Oriya handwritten databases provided by the ISI (Indian Statistical Institute, Kolkata)\footnote{http://www.isical.ac.in/~ujjwal/download/database.html}, CVL usually used for pattern recognition competitions\!\!\footnote{http://www.caa.tuwien.ac.at/cvl/category/research/cvl-databases/}, are employed in our experimental study as a benchmark resource for optical character recognition. Readers can refer to our previous work \cite{WangandLi-SCN} for more descriptions about these datasets. Here we summarize the basic information about these four databases in Table \ref{tab:5}. It should be noted that all the images of these four databases have been converted to grayscale and normalized size ($28\times28$) by applying the similar preprocessing procedures used in MNIST database \footnote{http://yann.lecun.com/exdb/mnist/}, that is, scaled to 20x20 by preserving the aspect ratio and then the centroid of the image is placed on the center of a standard plane of size 28x28. Some sample images are shown in Fig. \ref{fig:handwritten}.
\begin{table}[!h]
\caption{Statistics of handwritten digits databases}\label{tab:5}
\begin{center}
\begin{tabular}{cccc}
\toprule
Name&   Number of Images& Training &Test \\
\midrule
 Bangla&  23392& 19392 & 4000\\
 Devnagari& 22556& 18794&  3762\\
 Oriya& 5970& 4970 &1000\\
 CVL& 35780&  14000  &21780\\
\bottomrule
\end{tabular}
\end{center}
\end{table}
\begin{figure*}[htbp!]
\centering
\subfigure[Bangla:Training]{\includegraphics[width=0.245\textwidth]{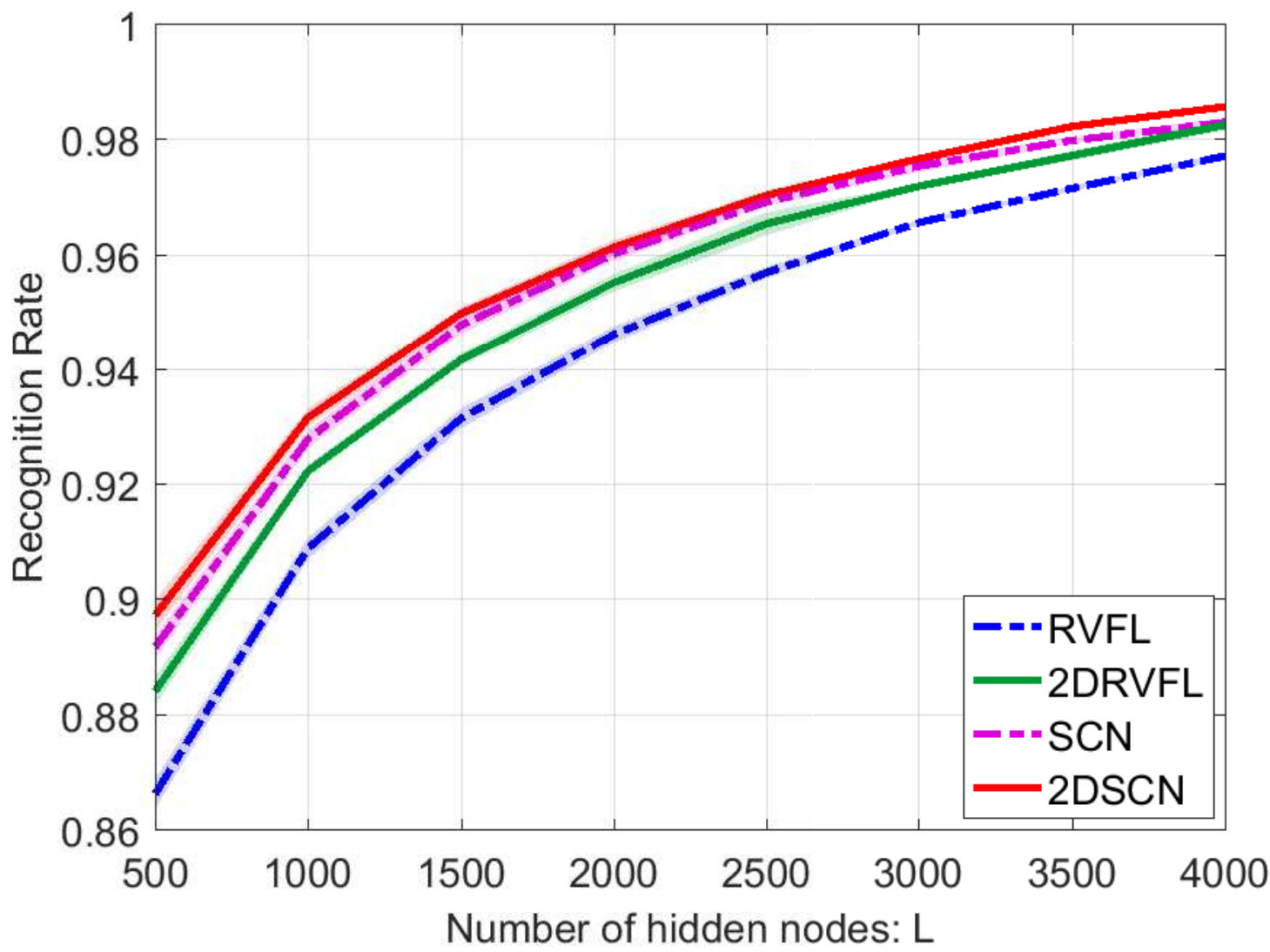}}
\subfigure[Devnagari:Training]{\includegraphics[width=0.245\textwidth]{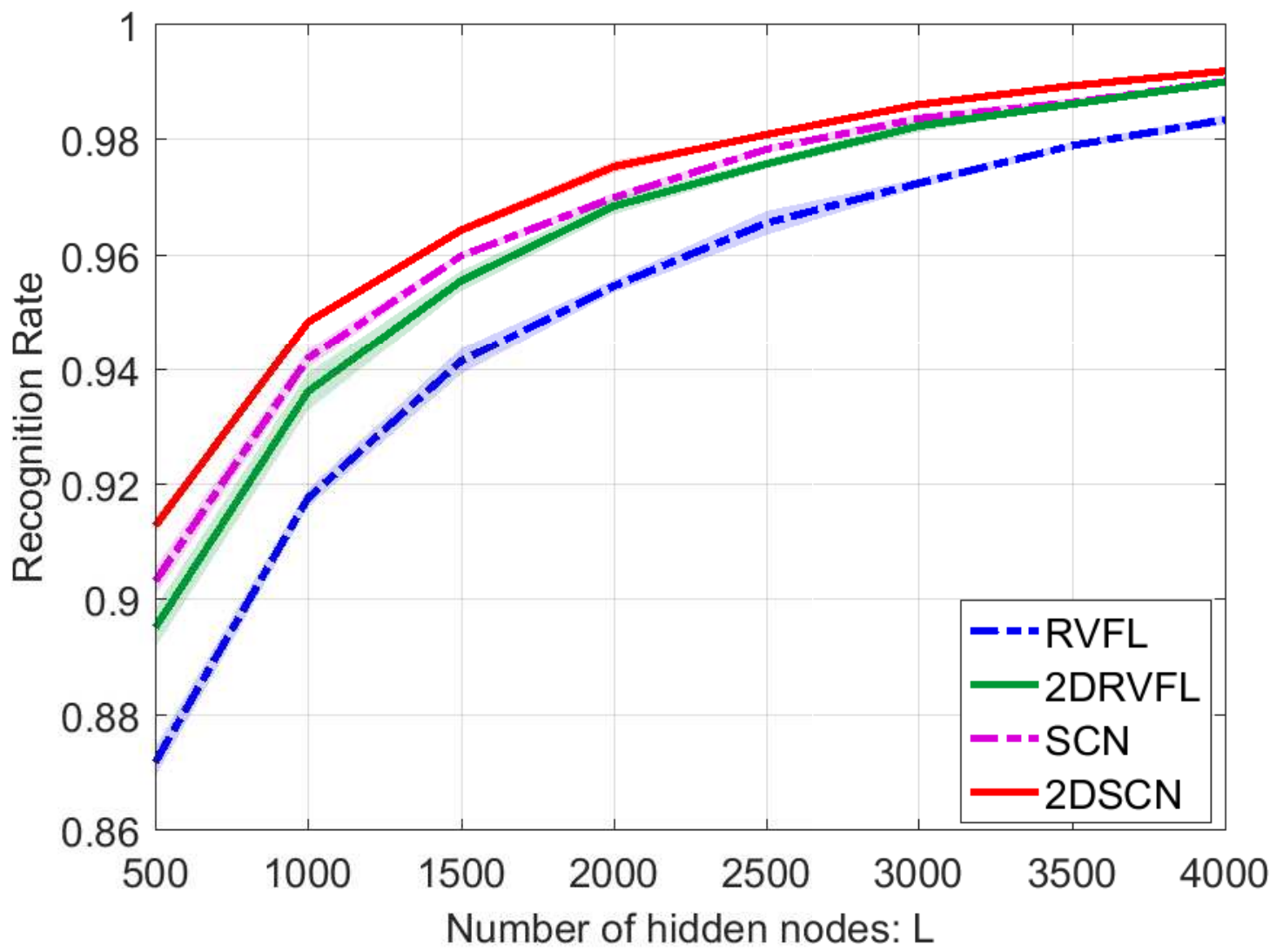}}
\subfigure[Oriya:Training]{\includegraphics[width=0.245\textwidth]{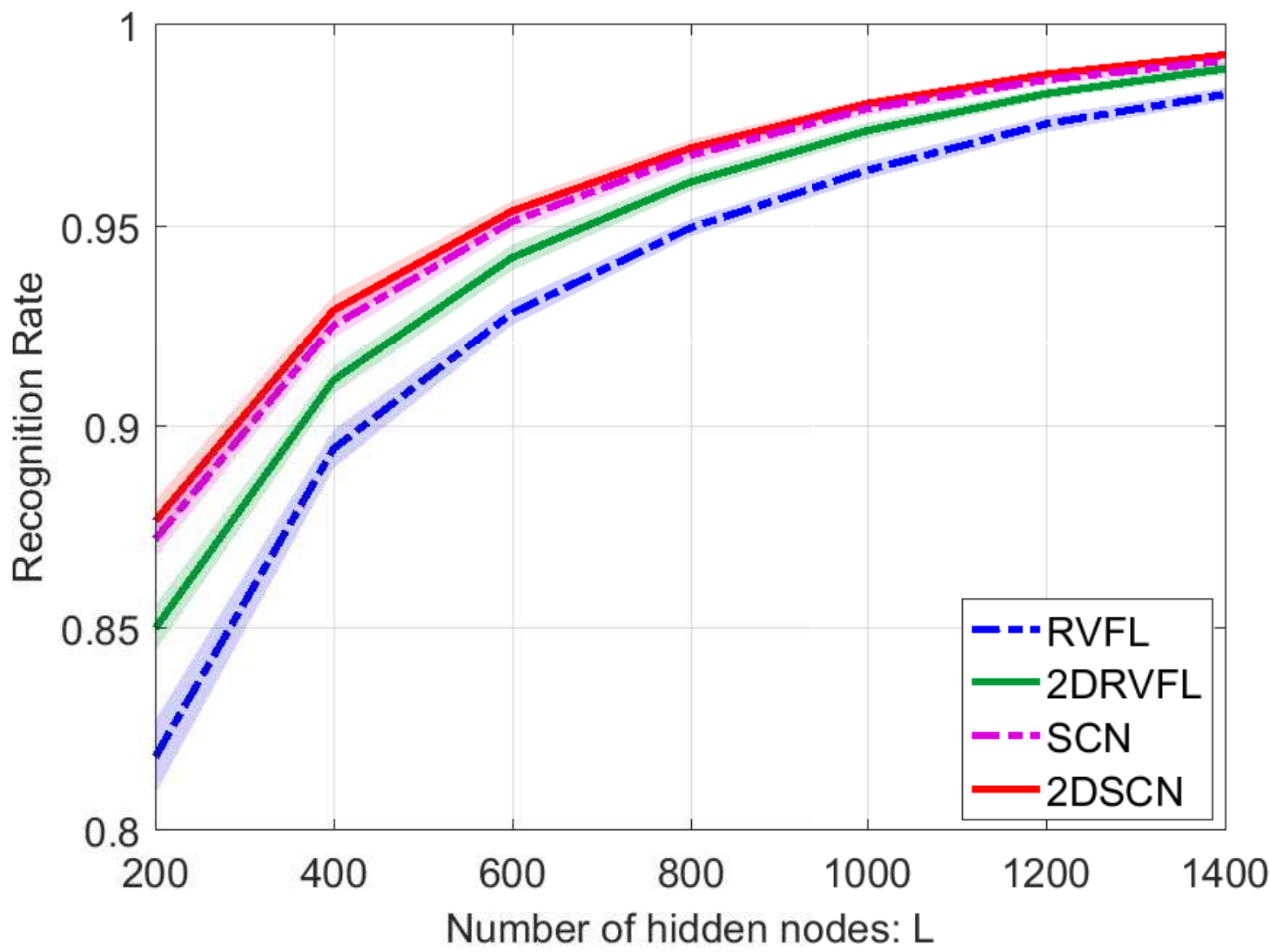}}
\subfigure[CVL:Training]{\includegraphics[width=0.245\textwidth]{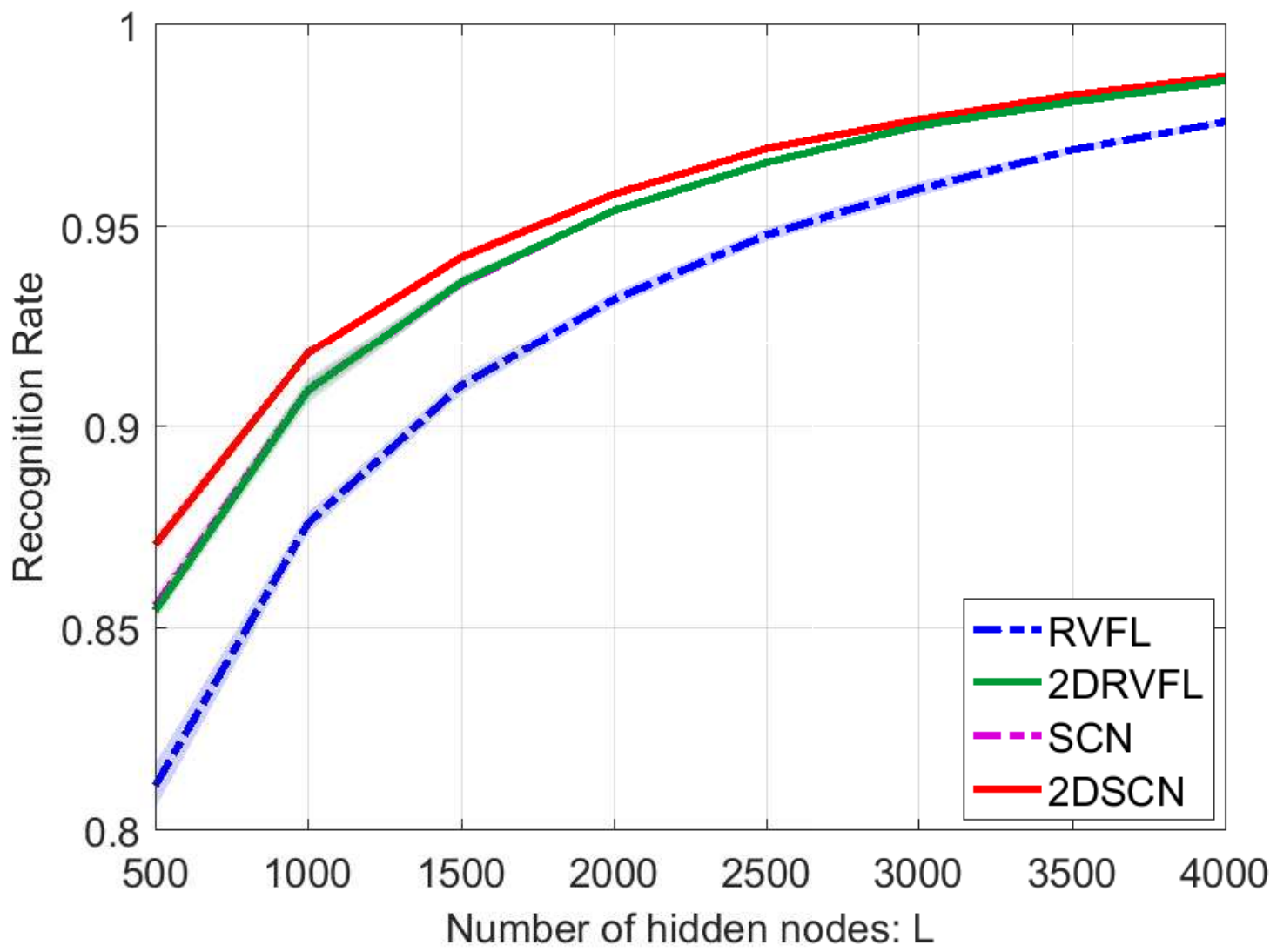}}
\subfigure[Bangla:Test]{\includegraphics[width=0.245\textwidth]{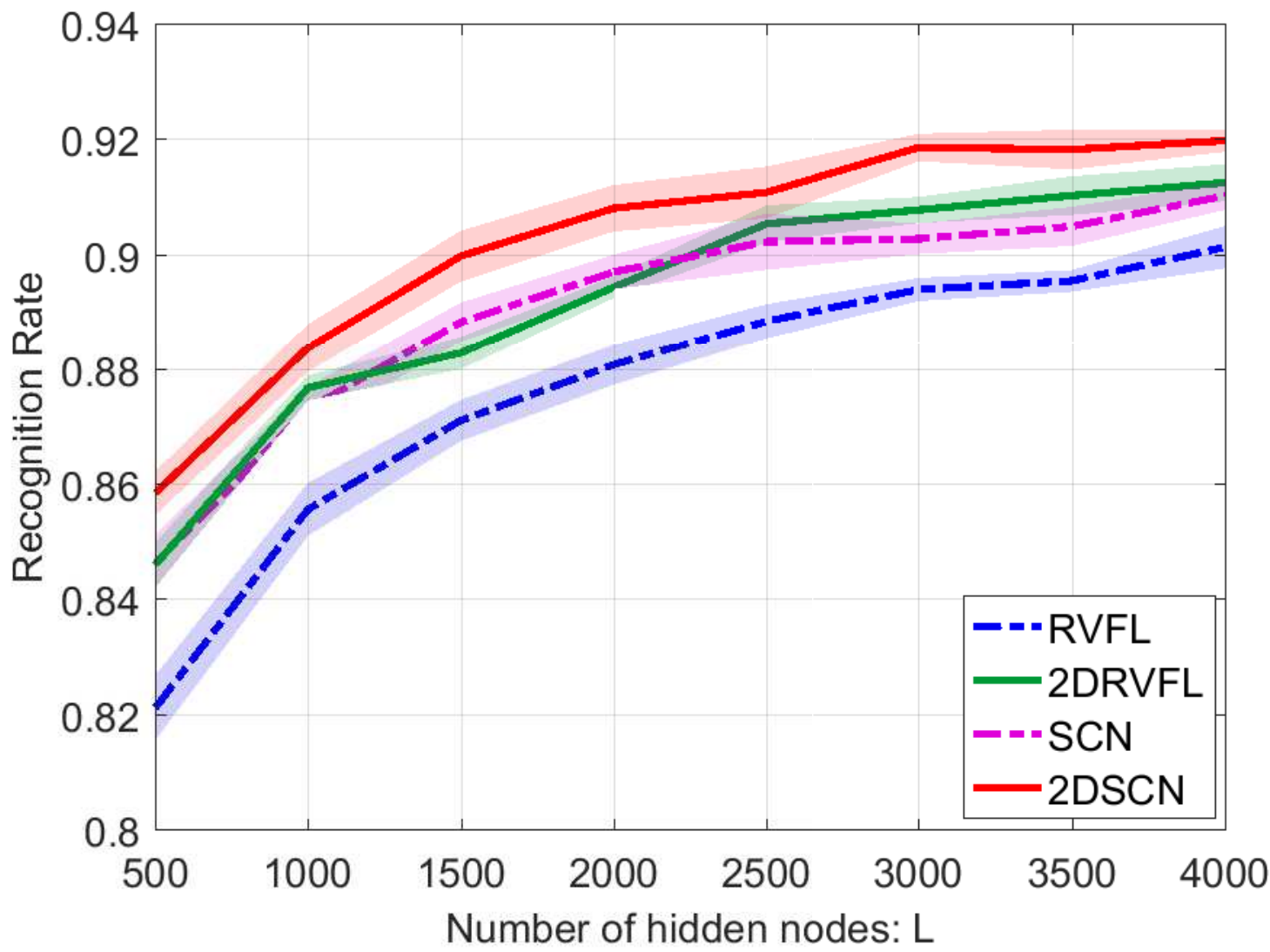}}
\subfigure[Devnagari:Test]{\includegraphics[width=0.245\textwidth]{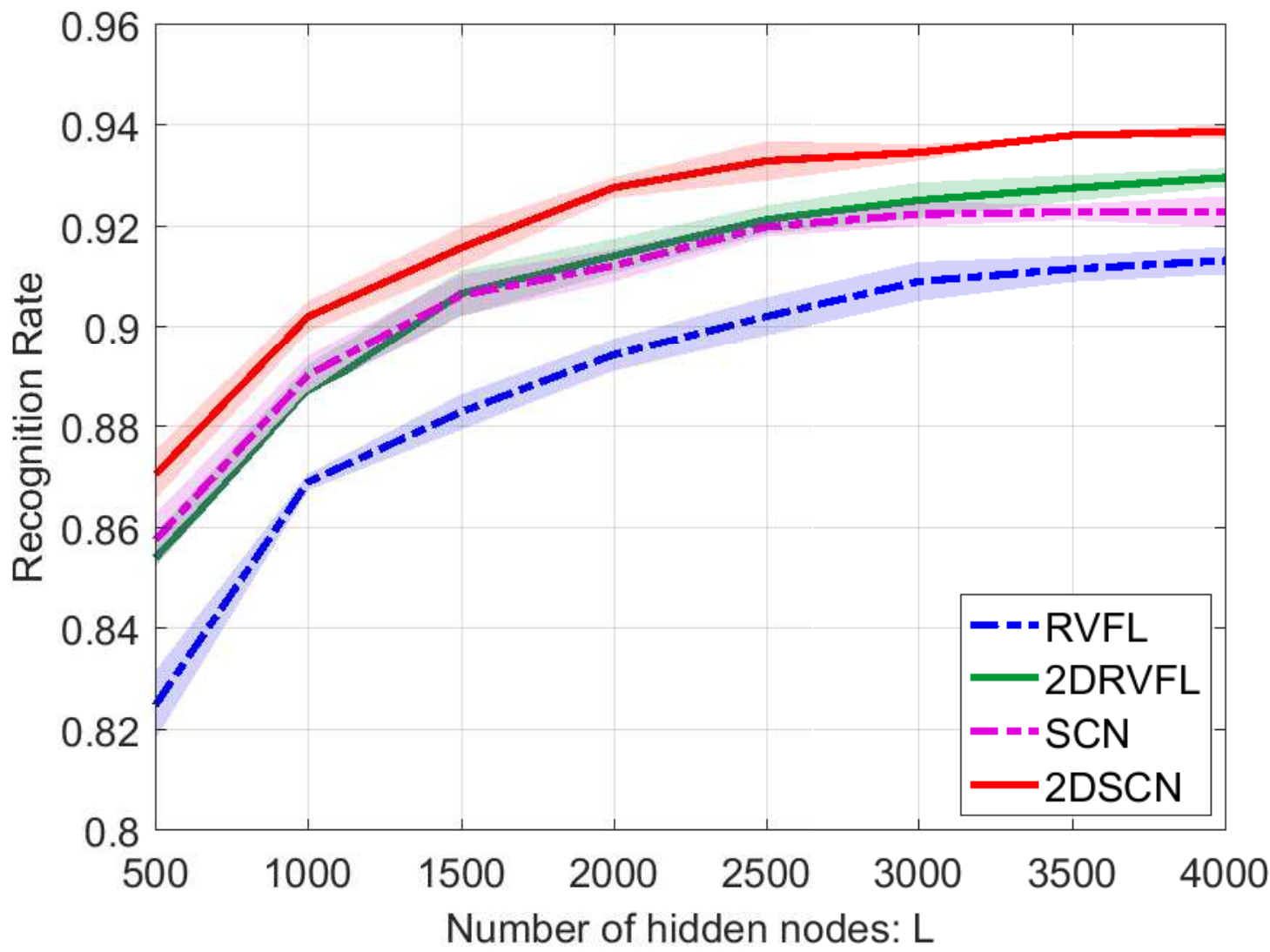}}
\subfigure[Oriya:Test]{\includegraphics[width=0.245\textwidth]{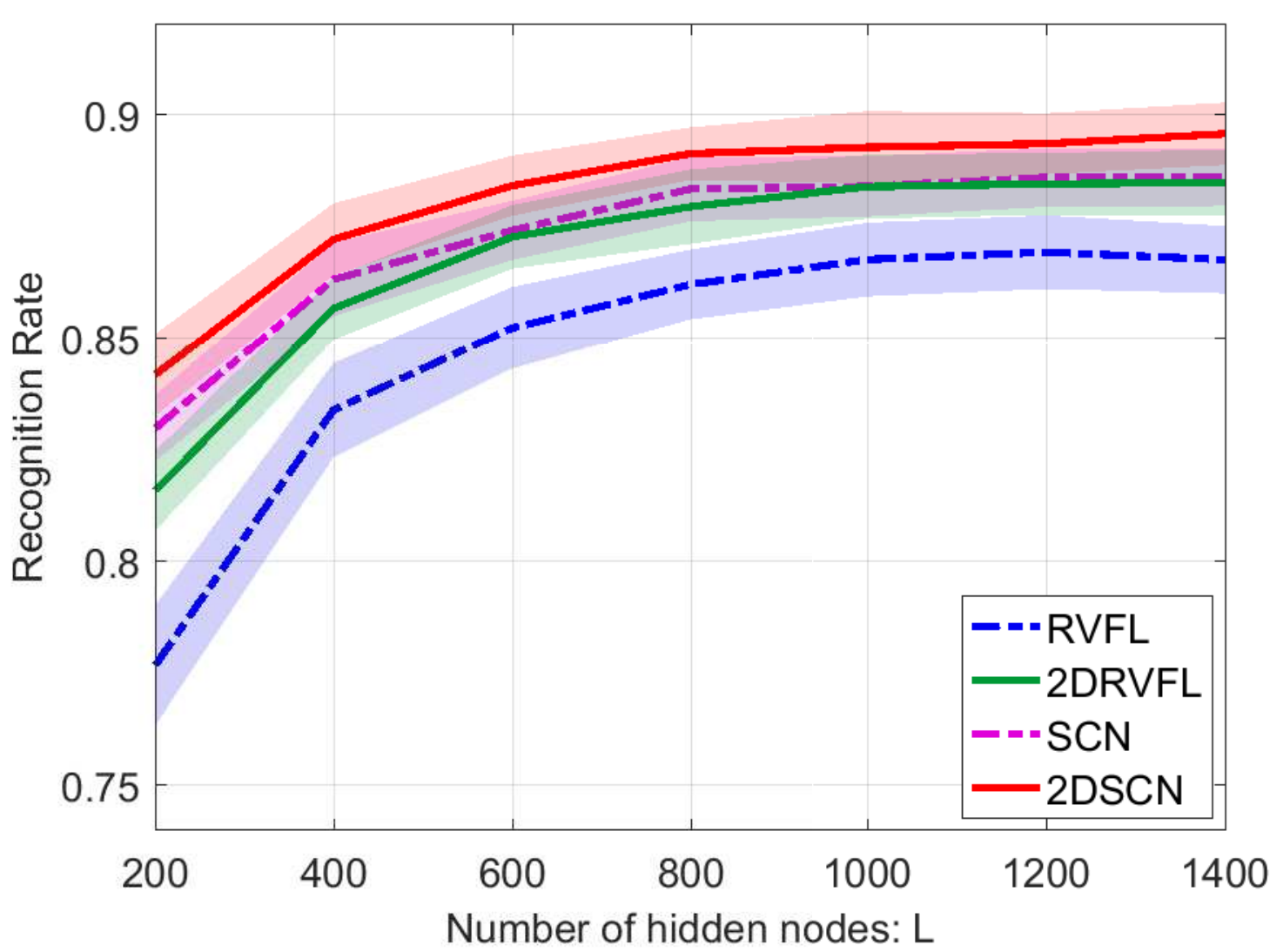}}
\subfigure[CVL:Test]{\includegraphics[width=0.245\textwidth]{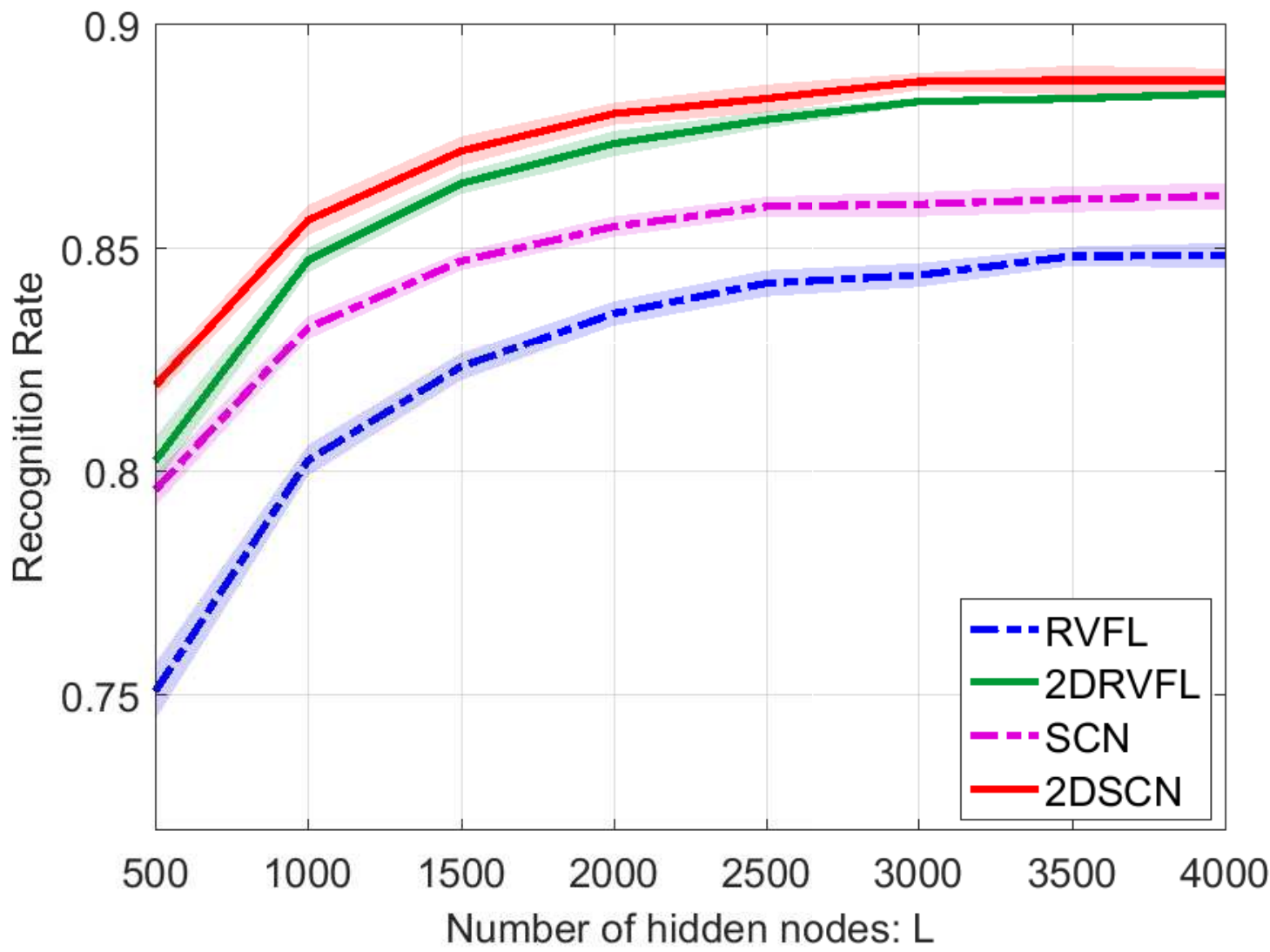}}
\caption{Performance comparison among 2DSCN, 2DRVFL, SCN, and RVFL on Bangla, Devnagari, Oriya, and CVL database.}\label{fig:recognition comparison}
\end{figure*}

\subsubsection{Results and Discussion}
Fig. 5 displays performance comparison for both training and test in terms of different setting of $L$, in which the mean and standard deviation values of recognition rate are based on 10 independent trials (More trails can indicate more convinced results but not necessary based on our experience, because the  standard deviation values are relatively small and stable at certain level). It is apparent that 2DSCN outperforms the others while RVFL has the worst results in all these four datasets. Interestingly, the test results of 2DSCN are much better than that of SCN, even when their training recognition rate is relatively close. Similar finding can be noticed among the comparison between 2DRVFL and SCN. For example, for Bangla, Devnagari, and Oriya, 2DRVFL occasionally can result in slightly higher recognition rates in testing even the corresponding training recognition rate is lower/quite close to that of SCN. It becomes much more clear in the subplots for CVL dataset, i.e., 2DRVFL works more favorably than SCN but still worse than 2DSCN, which to some extent lends strong support for our theoretical investigation on the superiority of randomized learners with 2D inputs.

\subsection{Case Study: Human Face Recognition}
In this section, we further demonstrate the advantage of 2DSCN over the other three randomized learner models on human face recognition tasks, where the input dimensionality is far more larger than that of the handwritten digit problems addressed before. Followed by a brief description of the used databases, we present the performance comparison for these four methods. Later, as a verification for Theorem 5, we calculate the corresponding test error upper bounds for the randomized models produced by those four algorithms, and visualize their differences for both datasets. In addition, we compare the capabilities of those four algorithms in dealing with training images corrupted by certain level of random noises, aiming to show their robustness in problem-solving. Generally, our experimental results can strongly support the superiority of 2DSCN algorithm on image data modelling problems.

\subsubsection{Databases}
\begin{figure}[htbp!]
\centering
\subfigure[ORL]{\includegraphics[width=0.48\textwidth]{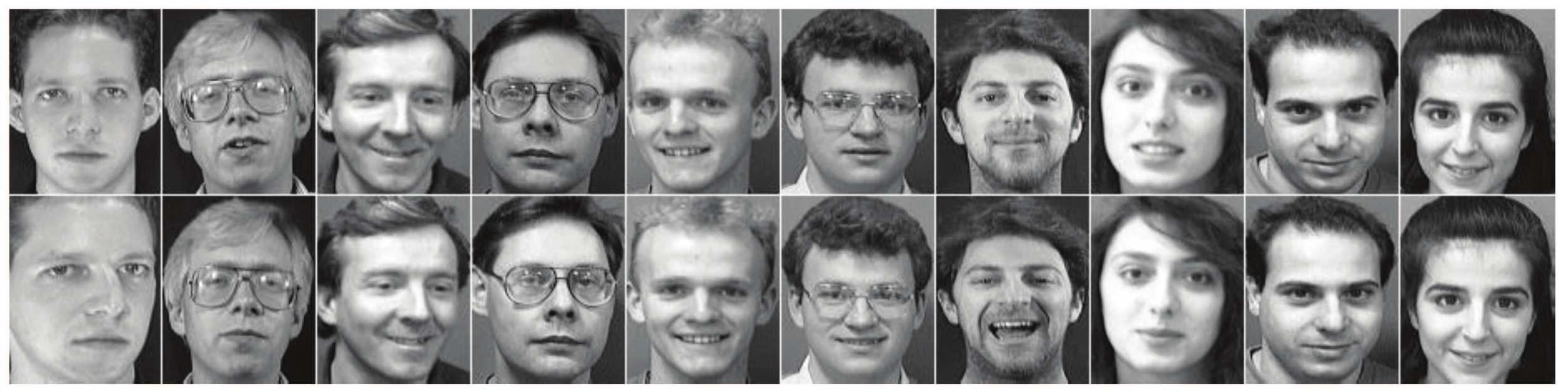}}
\subfigure[FERET]{\includegraphics[width=0.48\textwidth]{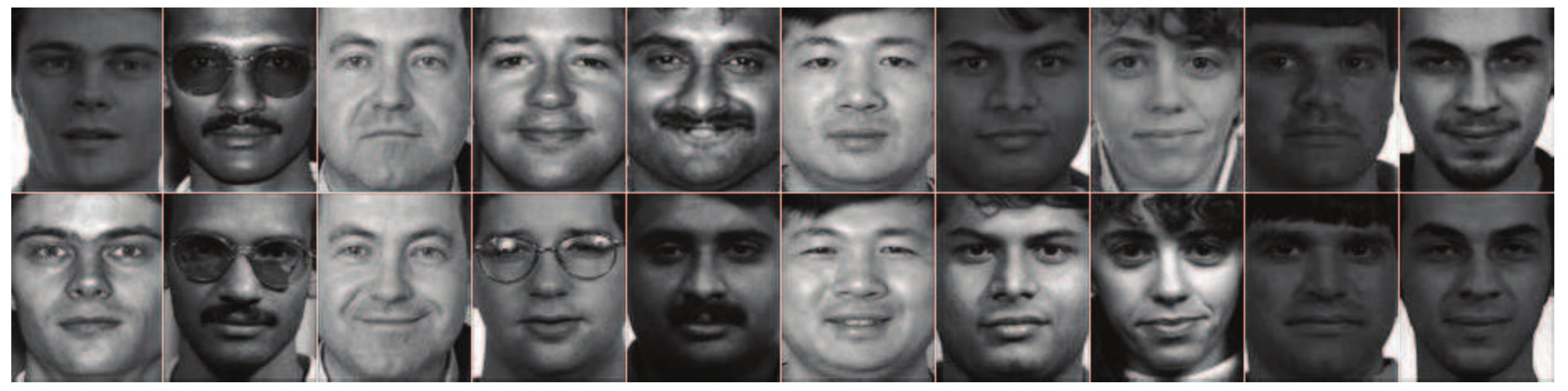}}
\caption{Samples of face images from ORL, FERET database. Ten subjects with two expressions for each are displayed.}
\end{figure}
Two benchmark datasets for human face recognition problem are employed in our experiments, as we introduce in the following.
\emph{ORL} \cite{ORL}: The Olivetti (ORL, now AT$\&$T) database contains ten $112\times 92$ pixel gray scale images of 40 distinct subjects (individuals). Among those subjects, some images were taken at different times, with various lighting conditions, different types of facial expressions (open/closed eyes, smiling/not smiling) and varying facial details (glasses/no glasses). Specifically, all the images were taken against a dark homogeneous background with the subjects in an upright, frontal position (with tolerance for some side movement).

\emph{FERET} \cite{FERET}: The Facial Recognition Technology (FERET) database contains a total of 14,126 gray scale images for 1199 subjects, which were collected over several sessions spanning over three years. In particular, for some individuals, over two years had elapsed between their first and last sittings, with some subjects being photographed multiple times. In our experimental study, we choose 72 subjects with 6 frontal images of size $112\times 92$ per person.

For these two databases, we randomly select half of the images for each subject as the training samples and the other half for test.

\subsubsection{Results and Discussion}
\begin{figure}[htbp!]
\centering
\subfigure[ORL]{\includegraphics[width=0.24\textwidth]{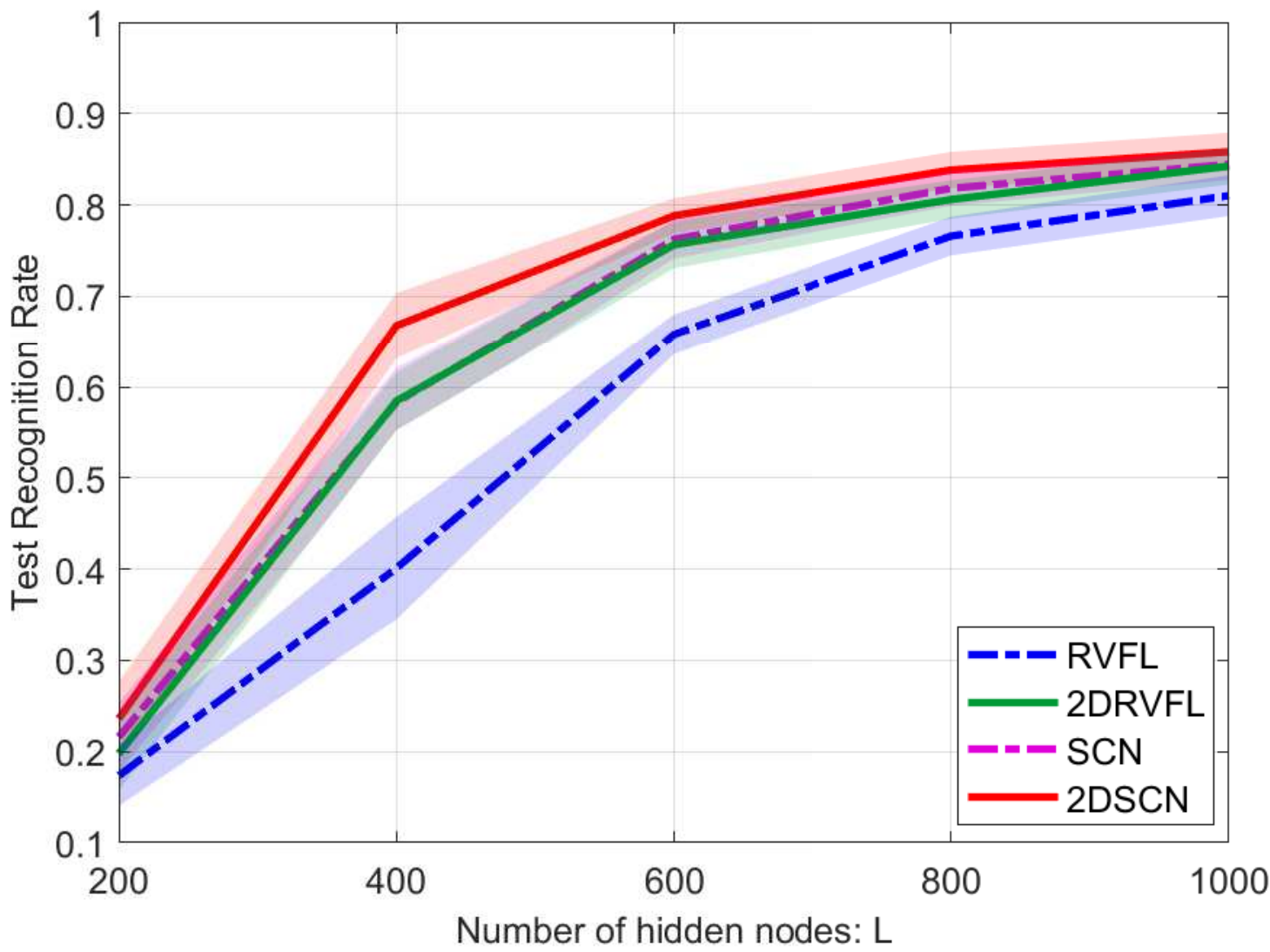}}
\subfigure[FERET]{\includegraphics[width=0.24\textwidth]{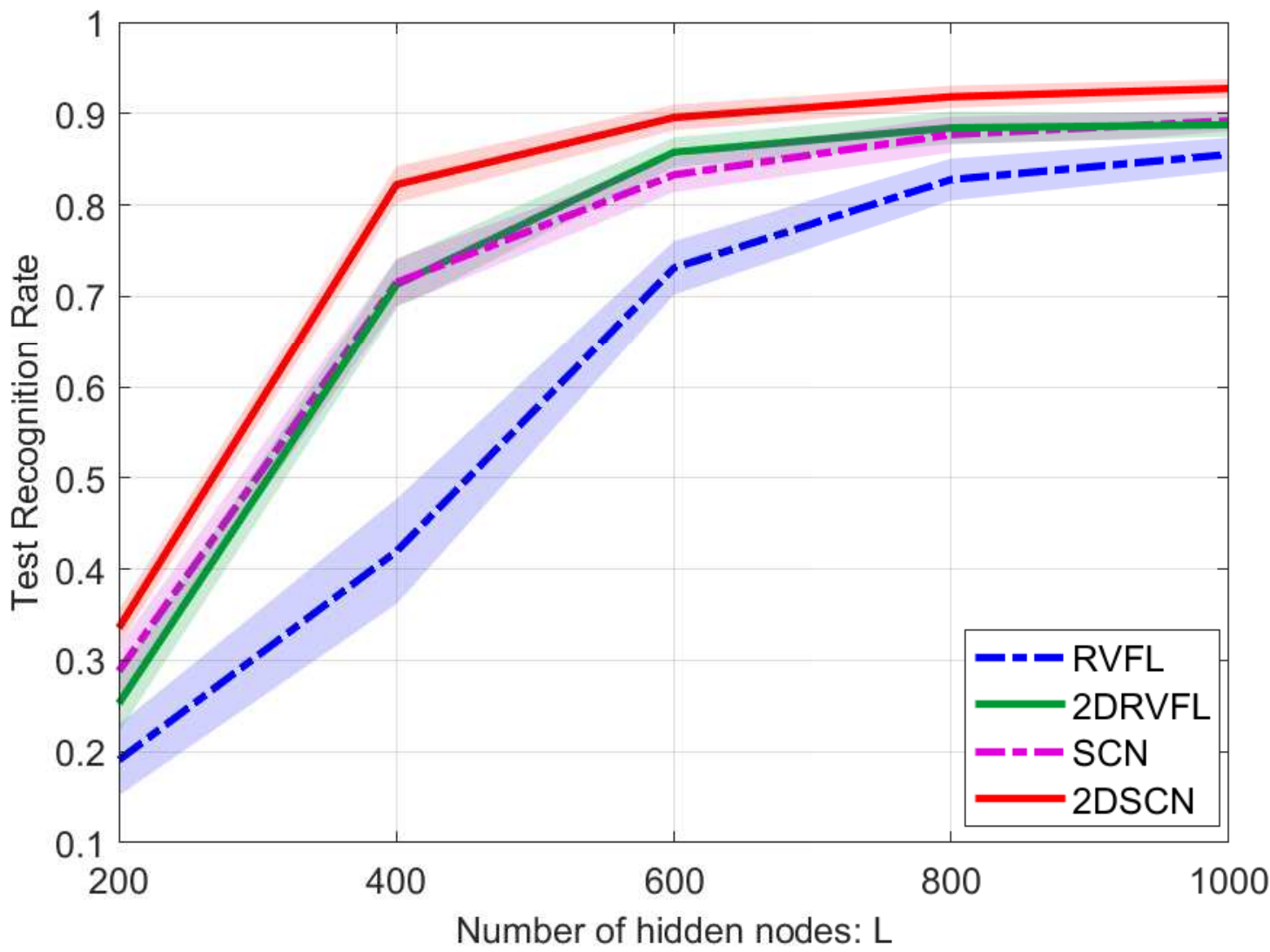}}
\caption{Test recognition rate comparison for 2DSCN, 2DRVFL, SCN, and RVFL on ORL and FERET.}\label{HumanFacePerformance}
\end{figure}

In our experiments, we employ the same parameter setting for these four algorithms (see part A for details), and conduct the performance comparison with various setup for the number of hidden nodes, i.e., $L=200,400,600,800,1000$, respectively. For each dataset, 50 independent trials are performed on each $L$ for all four algorithms.  Fig. \ref{HumanFacePerformance} shows their test recognition rates with both mean and standard deviation values. Clearly and similar to what we have demonstrated before, 2DSCN well outperforms the other three algorithms in generalization. On the other hand, we should note that the training recognition rate for all these algorithms in all cases is 1, which means that each method with the current parameter setting exhibit sufficient learning capability, however, has significant discrepancy in generalization ability. Besides, as shown in \ref{HumanFacePerformance}, there is no big difference between the performance of 2DRVFL and that of SCN, and both of them contribute to a better result than RVFL. Now, it is fair to say that 2D models have more advantages than 1D models in both training and test, such as what we have observed in previous Part A and B.
Based on Theorem 5 in Section III, we can numerically estimate the test error upper bound for the learner model produced by each of these four algorithms. Note that the upper bound in Eq. (\ref{theorem4}) contain three parts, that is, training error $\|H\beta-Y\|_{F}$, $\epsilon\max_{1\leq i \leq N}\|Z_i\|_2\!\cdot\!\|H\!\circ\!(O\!-\!H)\!\circ\!\ddot{W}\|_{F}\|\beta\|_{F}$, and the sufficiently small term $o(\epsilon^2)\|\beta\|_{F}$, respectively.
Since their training errors stay in the same level (training recognition rate equals to 1 for each case) and the fact that $\|\beta\|_{F}$ also occurs in the second part, we only need to consider the second term but without the common factor $\max_{1\leq i \leq N}\|Z_i\|_2$ in our empirical examination. In particular, for each method, we consider the case of $L=600$ and calculate the value of $\|H\circ(O-H)\circ\ddot{W}\|_{F}\cdot\|\beta\|_{F}$ for 50 resulted leaner models based on independent trials, followed by a normalization operation via dividing the corresponding figure by the maximum value of all the  $50\times 4=200$ records. We call this numerical (predictive) upper bound the generalization indicator $\Theta:=\{\Theta_i\}_{i=1}^{200}$, denoted by (refer to Section III for some notations)
\begin{equation*}
\Theta_i=\frac{\{\|H\circ(O-H)\circ\ddot{W}\|_{F}\cdot\|\beta\|_{F}\}_i}{\max\{\|H\circ(O-H)\circ\ddot{W}\|_{F}\cdot\|\beta\|_{F}\}_{i=1}^{200}},
\end{equation*}
where the index $i$ corresponds to the $i$-th record among the total 200 records.

\begin{figure}[htbp!]
\centering
\subfigure[ORL]{\includegraphics[width=0.24\textwidth]{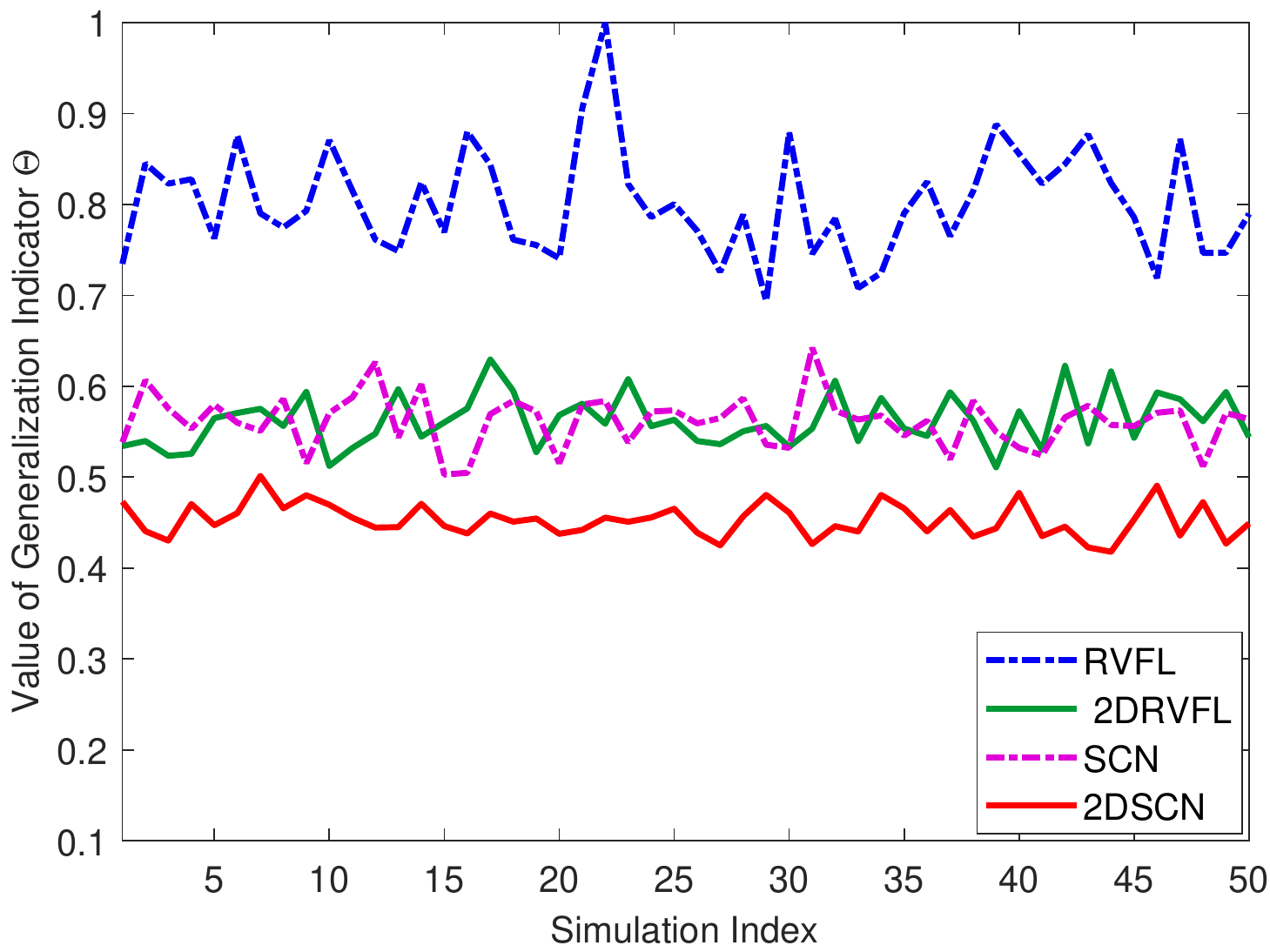}}
\subfigure[FERET]{\includegraphics[width=0.24\textwidth]{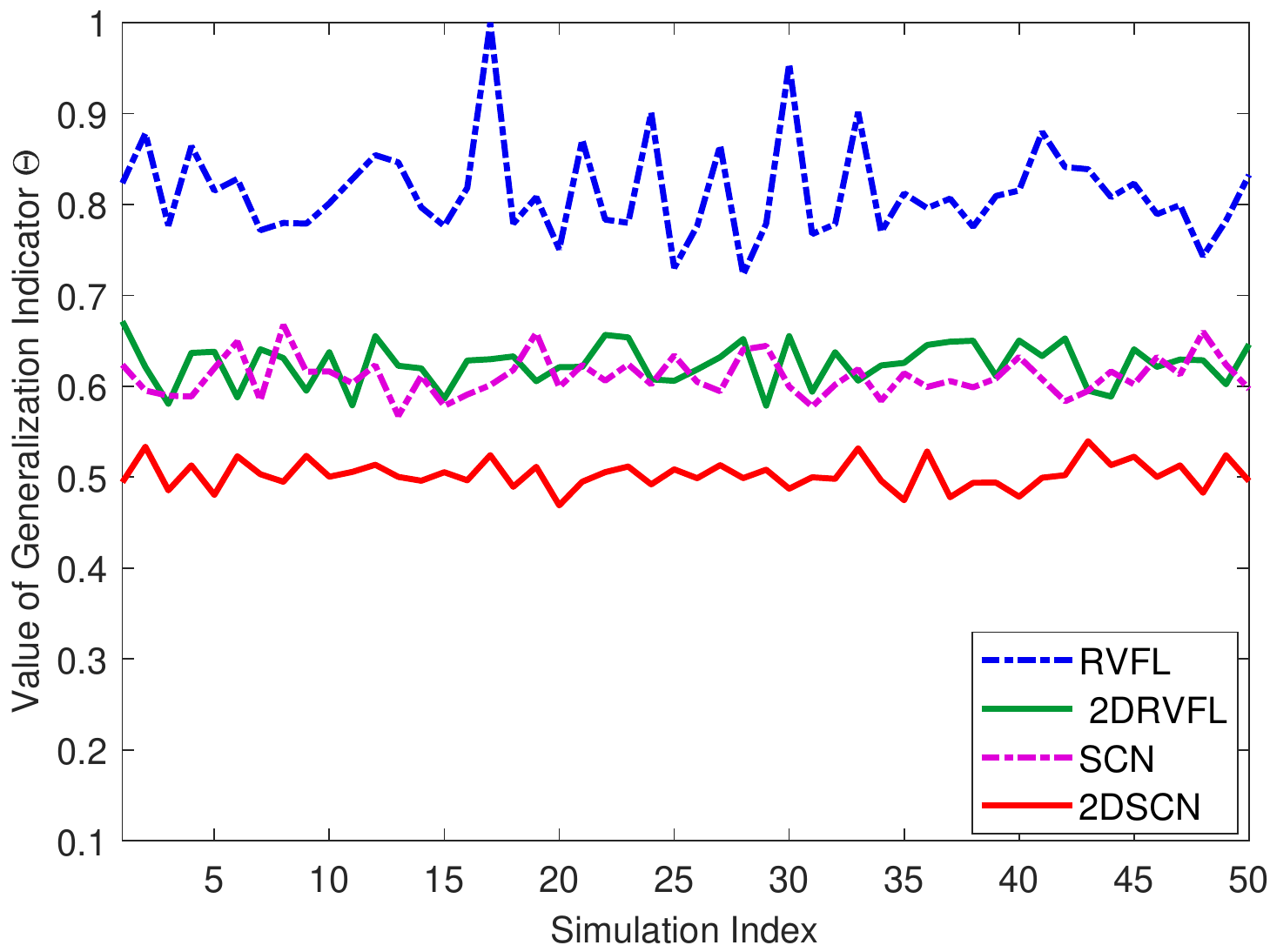}}
\caption{Test error upper bound comparison for 2DSCN, 2DRVFL, SCN, and RVFL on ORL and FERET.}\label{generalization_indicator}
\end{figure}

Fig. \ref{generalization_indicator} plots all these 200 records for the four algorithms (50 records for each), in which we can clearly observe that 2DSCN exhibits a lowest test error upper bound than the other three methods while RVFL has the highest results. Apparently and interestingly, this is consistent with their real test recognition rate comparison shown in Fig. \ref{HumanFacePerformance}, therefore, verify the effectiveness and practicability of our Theorem 5. Based on our experience, similar results can be obtained with the other option of $L$ setting, and as such, learner models built by 2DSCN algorithm have the smallest predictive test error upper bound estimation. As for the space limitation, more statistical results and analysis are left for our future work.
\subsubsection{Robustness Illustration}
\begin{figure}[htbp!]
\centering
\includegraphics[width=0.48\textwidth]{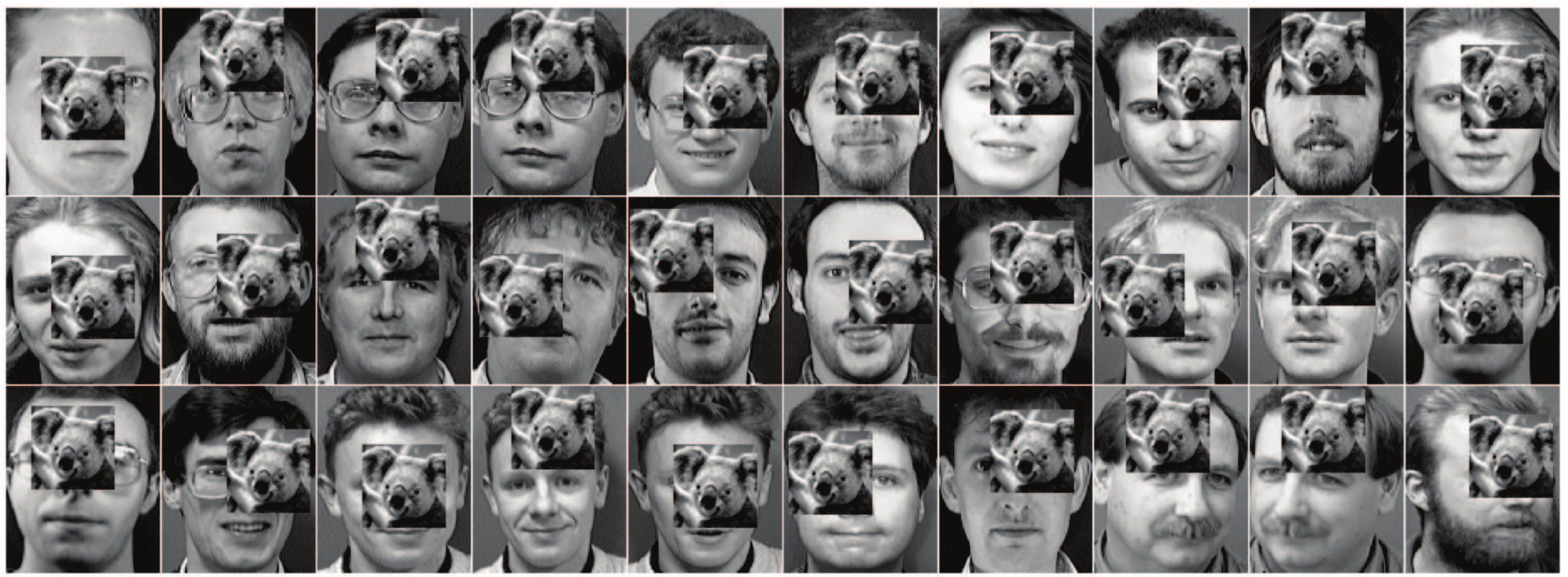}
\caption{Sample images from ORL with random block occlusion.}\label{outliers}
\end{figure}

To further investigate the superiority of 2DSCN over the other methods, we randomly select 50 images from ORL training set and artificially simulate contiguous occlusion by replacing a randomly located square block of each chosen image with an unrelated image (e.g., \emph{Koala}). 30 of these corrupted images are display in Fig. \ref{outliers}. Our objective is to compare the test performance of these four algorithms and study their capability in performing robustness in training, that is, to what extent they can alleviate the impacts of random block occlusion attached in the training images. All details for experimental setup remain the same as what we have used for the original (clean) ORL dataset.
\begin{figure}[htbp!]
\centering
\includegraphics[width=0.42\textwidth]{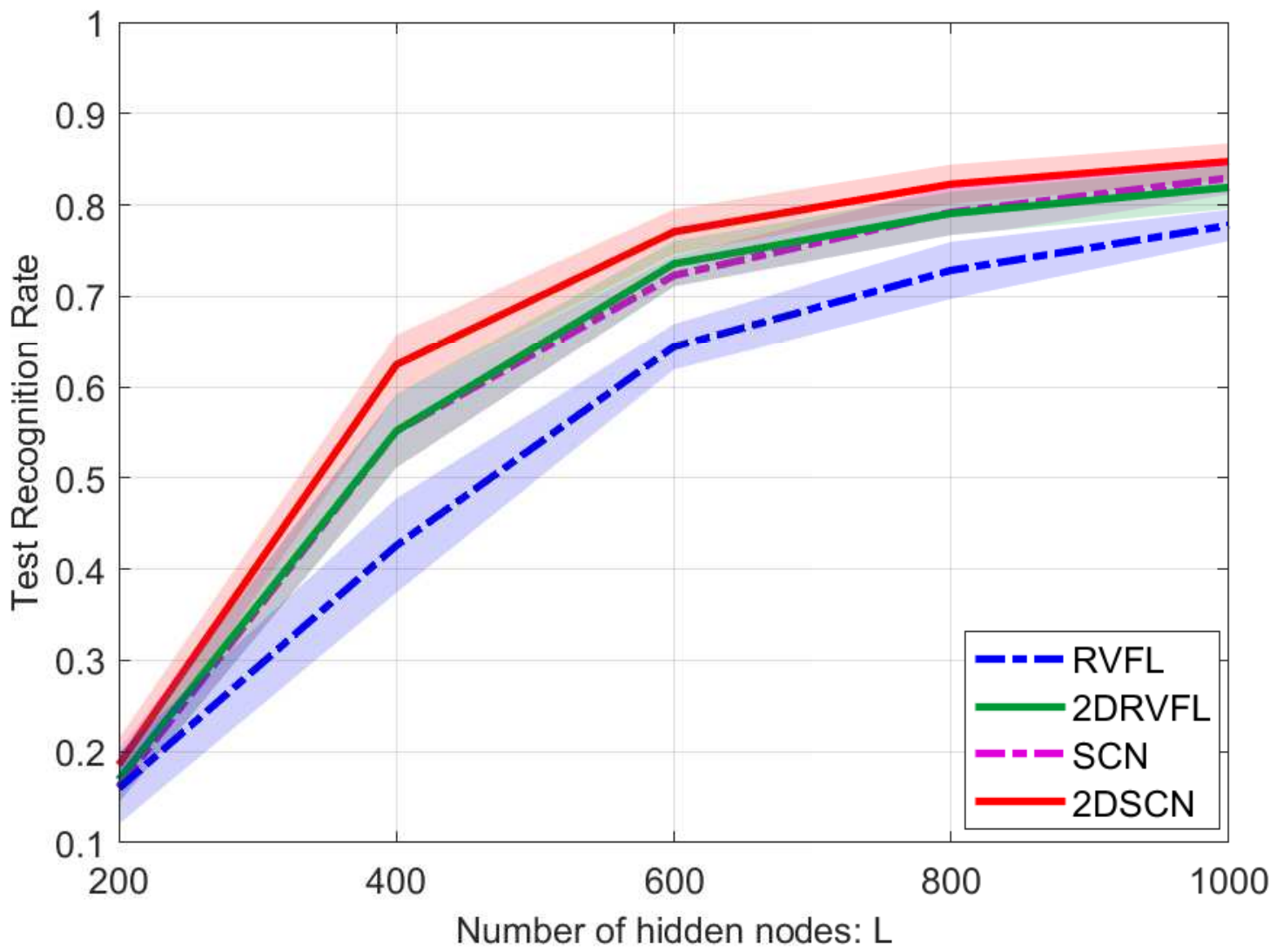}
\caption{Test recognition rate comparison for 2DSCN, 2DRVFL, SCN, and RVFL on corrupted ORL.}\label{outliers_performance}
\end{figure}

As can be seen in Fig. \ref{outliers_performance}, 2DSCN still performs best in generalization, at the same time, RVFL has the lowest test recognition rates. This observation can to some extent imply that 2D models may still exhibit their merits for problem-solving of robust image data modelling, on the basis of their underlying advantages as shown throughout this paper. We expect more interesting work employing/extending the current 2DSCN framework to deal with robust image recognition tasks or similar applications in parallel. By saying that, we have to admit that the technical issue around is out of our main focus in this work and more interesting follow-up researches are desirable in the future.

\section{Conclusions}
This paper develops two dimensional stochastic configuration networks (2DSCNs), which extend the original SCN framework for data analytics with matrix inputs. Compared to existing randomized learning techniques, the proposed algorithm maintains all the advantages of the original learning techniques for SCNs, such as fast modelling, universal approximation property, and sound generalization power. Some associations and differences between 2DSCNs and SCNs are theoretically investigated and empirically justified.  Our main technical contribution in this paper lies in an interpretation on reasons behind the improved generalization of 2DSCNs against the original SCNs for image data modelling. Compared to the performance obtained from SCNs, RVFLs and 2DRVFLs, we conclude that 2DSCNs outperform in terms of both learning and generalization, and have great potential for real world applications.  

There are many interesting studies left for future work. For example, a trivial extension of the current 2DSCN to robust version can be realized by an immediate combination of this work and our previous research \cite{wang2017robust}. It is also important to point out that, while this work focuses on shallow neural networks, the framework and the associated theoretical analysis are generic and general enough to adopt the deep machinery, therefore, one can make efforts towards building a deep 2DSCN to maintain both the superiority of DeepSCN \cite{WangandLi-DeepSCN} and the advantages of 2DSCN. It is also of practical importance to employ the proposed 2DSCN in stream image data modelling. Other than than, it would be interesting to further enhance 2DSCN by considering the regularization learning framework, or refined stochastic configuration inequality with sparsity constrains.


%
%

\ifCLASSOPTIONcaptionsoff
  \newpage
\fi

\end{document}